\documentclass[10pt]{article}
\usepackage[accepted]{tmlr}

\usepackage{amsmath,amsfonts,bm}

\def\eqref#1{equation~\ref{#1}}

\def\1{\bm{1}}

\DeclareMathAlphabet{\mathsfit}{\encodingdefault}{\sfdefault}{m}{sl}
\SetMathAlphabet{\mathsfit}{bold}{\encodingdefault}{\sfdefault}{bx}{n}

\usepackage{hyperref}
\usepackage{url}
\usepackage{graphicx}
\usepackage{booktabs}
\usepackage{adjustbox}
\usepackage[table]{xcolor}
\usepackage{subcaption}
\usepackage{enumitem}
\usepackage{amssymb}
\usepackage{float}
\usepackage{longtable}

\newtheorem{definition}{Definition}
\newtheorem{theorem}{Theorem}

\title{{Evaluating Ensemble Methods for Algorithmic Stability \\[0.1em] Under Data Perturbations}}
 
\author{\name Benjamin Edelhertz \email benedel@uw.edu \\
      \addr Department of Statistics\\
      University of Washington
      \AND
      \name Simon D. Nguyen \email simondn@uw.edu\\
      \addr Department of Statistics \\
      University of Washington
      \AND
      \name Tyler H. McCormick \email tylermc@uw.edu\\
      \addr Department of Statistics\\ Department of Sociology\\
      University of Washington}

\def\month{11}
\def\year{2026}
\def\openreview{\url{https://openreview.net/forum?id=7CCr0dgNwm}}

\begin{document}

\maketitle

\begin{abstract}
 Standard classification algorithms often exhibit sensitivity near decision boundaries, which can compound instability that already exists from overfitting to a subset of high-influence samples. Ensembling is commonly used to stabilize classifiers, but we show that it can destabilize those that are already stable. The well-established practice of subbagging can yield substantial improvements in stability, as we validate in the cause-of-death classification setting. However, its effect depends on the perturbation level, and for stable classifiers the finite number of bags introduces destabilizing Monte Carlo noise that shrinks only slowly as the number of bags grows. This instability can be mitigated, though, by the recently proposed inflated argmax, which allows for multi-class predictions in ambiguous cases. We further examine the extent to which the Rashomon set --- the collection of distinct models that achieve near-optimal predictive performance --- can reduce instability by leveraging predictive multiplicity rather than resampling. We find that the Rashomon ensemble can stabilize unstable classifiers, but requires sufficient predictive multiplicity, which is not guaranteed. It is a promising direction, but not currently a general substitute for subbagging. Our results are consistent with an influence equalization interpretation of why these ensemble approaches stabilize unstable classifiers, although their failures on stable classifiers appear to have distinct sources.
\end{abstract}

\section{Introduction}

Stability is the degree to which the outputs, or statistical conclusions more broadly, of an algorithm remain consistent under perturbations to the training data \citep{Yu_2013}. It shapes the level of trust we can place in the algorithm. When predictions are highly sensitive to the minute details of which data happened to be collected, then conclusions drawn from them risk reflecting noise in the sample rather than a real pattern in the world, undermining reproducibility and the fairness of subsequent decisions. The notion of stability has been well explored in the literature and has been shown to have significant downstream benefits in statistical analysis. \citet{learnability2010} pinpoint stability as a necessary and sufficient condition for learnability. Likewise, \citet{bousquet2002stability} and \citet{poggio2004general} relate stability to generalization, using stability to derive generalization error bounds and identifying stability as a sufficient condition for the generalization of learning algorithms. \citet{Yu_2013} shows that stability is important for interpretable and reliable scientific models and that it plays a key role in sensitivity analysis and effect detection. \citet{Murdoch_2019} go so far as to say that ``one should not interpret parts of a model which are not stable to appropriate perturbations to the model and data.''

This paper serves to evaluate and compare key methods that can help achieve algorithmic stability and to demonstrate why and under what conditions they succeed or fail. The mechanisms behind these methods can be hard to decipher, and the existing literature is especially sparse when it comes to the shortcomings of these methods. We aim to outline the potential consequences of adopting each technique and provide a unified account that helps practitioners decide which approach is most appropriate for a given setting, since there is no method that dominates in every scenario.

There are many ways to measure the stability --- or conversely the instability --- of a learning algorithm \citep{bousquet2002stability}, though a straightforward one is the difference between the predictions of the model trained on the full dataset and those of models trained on leave-one-out subsamples, which we refer to interchangeably as jackknife perturbations. Bootstrap and other resampling approaches can also work, depending on the desired level of data perturbation. The ensemble approach of subbagging, combined with the inflated argmax selection rule, has been shown to significantly reduce instability \citep{inflated_argmax}. 

According to the interpretation of \citet{grandvalet2004bagging}, subbagging provides a stabilization effect by equalizing the influence of training samples. The inflated argmax then smooths hard decision boundaries in the selection process, further stabilizing predictions but requiring multi-class predictions for ambiguous cases. We consider using the Rashomon ensemble as a novel alternative, which relies on the Rashomon set to build an ensemble. Despite relying on different mechanisms --- resampling versus predictive multiplicity --- both approaches appear to achieve stability through the same underlying influence equalization effect. However, as we will see, these methods do not work universally, as stable and unstable base algorithms react differently to ensembling.

Our findings are demonstrated primarily in the context of using verbal autopsy data for cause-of-death assignment, a setting where prediction consistency is critical. Verbal autopsy is a method used to determine the cause of death through interviews with the next of kin or caregivers of the deceased, using a standardized questionnaire to gather details on symptoms, medical history, and the circumstances leading to the death. Throughout much of the world, full autopsies are rare, and so this methodology has become increasingly important for assigning causes to otherwise uncertified deaths. Mortality is a continuous process, and the prominent causes of death shift over time as diseases or other risk factors emerge, decline, or spread across populations. Furthermore, as our understanding of new and even existing diseases progresses, such as with COVID-19, we get a clearer picture of the true underlying causes of death for certain symptoms, further altering the training data. Moreover, the models are trained and deployed across populations from different regions around the globe, making generalization a key component.

\section{Background}

There is a distinction to be made between traditional output stability and selection stability. The former deals with the real-valued output of the model --- i.e., predicted class probabilities --- while the latter deals with the predicted label, often determined by some selection criteria. These are related but often not the same, and output stability does not always imply selection stability \citep{inflated_argmax}. Both are worth considering, but in classification settings we usually ultimately care about selection stability as the predicted classes are the end result ultimately used to make decisions. In the multi-class prediction setting, it is important to understand the different ways we can define selection stability and how that choice will affect model selection.

\subsection{Definitions of Instability}

We consider a training dataset $\mathcal{D} = \big((X_i, Y_i)\big)_{i\in [n]}$ with $n$ observations of features $X \in \mathcal{X}$ and labels $Y\in[L] = \{1,\dots,L\}$. Borrowing notation from \citet{inflated_argmax}, we say that a classification algorithm $\mathcal{C}$ has selection stability $\delta$ at sample size $n$ if, for all datasets $\mathcal{D} \in (\mathcal{X}\times [L])^n$ and all test features $x\in \mathcal{X}$,
\[
\frac{1}{n}\sum_{i=1}^n \phi(\hat{S},\hat{S}^{\setminus i}) \le \delta,
\]
where $\hat{S}=\mathcal{C}(\mathcal{D}, x)$
and $\hat{S}^{\setminus i}=\mathcal{C}(\mathcal{D}^{\setminus i},x)$ are the prediction sets on the test feature vector $x \in \mathcal{X}$ for the classifier $\mathcal{C}$ trained on the full training dataset $\mathcal{D}$ and the leave-one-out dataset $\mathcal{D}^{\setminus i}$, respectively. Finally, $\phi: 2^{[L]} \times 2^{[L]} \rightarrow [0,1]$ represents the instability metric as a function of the two prediction sets.

We focus on the mean and maximum instability across the samples in the test set, where the mean reflects typical instability and the maximum captures the worst-case behavior over the most ambiguous test points. A held-out test set is preferred, as we are primarily concerned with the stability of predictions on unseen data. Each test point will have a measure of instability calculated across the leave-one-out datasets. The instability for a single sample in the test set is the empirical mean of the instability score comparing the prediction sets of each leave-one-out model to the prediction set of the full model. Then the final measures for a given model are the mean and maximum instability across the test set. In this paper, we will use the following three definitions of instability:

\begin{equation}
    \delta_j^{\text{Disjoint}} := 1 - \frac{1}{n}\sum_{i=1}^n \mathbf{1}\left\{\hat{S}_j\cap \hat{S}_j^{\setminus i} \neq \varnothing\right\}
\end{equation}
\begin{equation}
    \delta_j^{\text{Overlap}} := 1 - \frac{1}{n} \sum_{i=1}^n \frac{|\hat{S}_j \cap \hat{S}_j^{\setminus i}|} {\min(|\hat{S}_j|, |\hat{S}_j^{\setminus i}|)}
\end{equation}
\begin{equation}
    \delta_j^{\text{Jaccard}} := 1 - \frac{1}{n} \sum_{i=1}^n \frac{|\hat{S}_j \cap \hat{S}_j^{\setminus i}|}{|\hat{S}_j \cup \hat{S}_j^{\setminus i}|}
\end{equation}

Here we denote $\delta_j$ as the instability for the $j$-th element of the test set. The term inside the summation is a stability score comparing two prediction sets, ranging from 0 (fully unstable) to 1 (fully stable). Each of these definitions is symmetric in $\hat{S}$ and $\hat{S}^{\setminus i}$ --- swapping the two prediction sets leaves the instability value unchanged --- which aids interpretation, since stability does not depend on an arbitrary choice of which set is treated as the reference. To say that a classification algorithm $\mathcal{C}$ has selection stability $\delta$ is to say that $\max_j(\delta_j) \leq \delta$.

The reason for outlining these similar, yet distinct, definitions is that each may favor a different type of model. The disjoint definition always favors a model that makes multi-class predictions, as the inflated argmax does, because it treats overlapping sets (i.e., sets that are neither equivalent nor disjoint) as fully stable. Thus, it does not capture the full extent of the similarity, or dissimilarity, of the prediction sets. The overlap and Jaccard definitions alter this to say that an equivalent set is more stable than an overlapping set, which is more stable than a disjoint set. Unlike the Jaccard definition, the overlap definition gives more leeway to different-sized sets, since the larger set need only contain each element of the smaller set in order to be fully stable.

Ultimately, the best definition is the one that fits the goals of the user, with specific consideration given to how reasonable and useful multi-class prediction sets are. In most cases, we recommend the overlap definition because it addresses the issue with the disjoint definition while not punishing dissimilarity solely due to differing set sizes. The reasoning is that the ordering of the predicted classes may be exactly the same, except that one more class made it into the prediction set, in which case we lean toward deeming that as generally stable. The set size is a separate measure that we would ideally minimize, so it makes sense to handle stability and set size somewhat separately. In practice, we found that the disjoint and overlap measures of stability rarely produce different results, while the Jaccard measure is noticeably more conservative.

\section{Related Work}

\subsection{The Inflated Argmax}

The obvious issue that arises specifically in classification problems is that decision boundaries are inherently unstable. When a model decides between multiple classes, it must implement a hard selection rule --- say taking the highest probability class. However, samples near the decision boundary are highly susceptible to being pushed over said boundary into a different class under perturbation. Hence, \citet{inflated_argmax} introduce the inflated argmax approach to address these issues that arise in classification settings. 

Intuitively, the inflated argmax works by softening the hard boundaries between classes. Instead of returning just the highest probability class as a singleton set, it returns the set of classes that could plausibly be the top class under some level of data perturbation. More specifically, it returns the set of all classes $j$ such that the predicted probability vector is within Euclidean distance $\varepsilon$ of some probability vector where class $j$ wins by a clear margin. The full definition from \citet{inflated_argmax} is as follows:

\begin{definition}[Inflated Argmax]
For any $w\in\mathbb{R}^L$, the \textit{inflated argmax} is defined as
\begin{equation}\label{eqn:define-stable-argmax}
\mathrm{argmax}^{\varepsilon}(w) := \left\{j\in[L] : \mathrm{dist}(w,R_j^{\varepsilon}) < \varepsilon\right\},
\end{equation}
where $\mathrm{dist}(w,R_j^\varepsilon) = \inf_{v\in R_j^\varepsilon}\|w-v\|$, and where
\[R_j^\varepsilon = \left\{w\in\mathbb{R}^L : w_j \geq \max_{\ell\neq j}w_\ell + \varepsilon/\sqrt{2}\right\}.\]
\end{definition}

By hedging our bets, in a sense, on ambiguous samples, we increase stability by avoiding uncertain predictions. With the standard argmax, small perturbations can push ambiguous samples over hard decision boundaries. With the inflated argmax, this no longer changes the prediction set because the classes on either side of the boundary are already included in the set, thus making the predictions stable. This is where the definition of instability becomes very important, since we are now allowing for multi-class predictions.

In their experiments, \citet{inflated_argmax} demonstrate that the inflated argmax approach can drastically increase stability. The caveat is that, in order to achieve these gains, it must be applied in concert with a subbagging ensemble procedure.

\subsection{Resampling Ensembles}

There are two primary resampling approaches that are similar in their effect but somewhat distinct in implementation: bagging and subbagging. Each approach resamples the data and refits the model on that data to generate predictions on a test set, then averages over those predictions.

With training data of sample size $n$:

\begin{itemize}
    \item \textbf{Bagging (Bootstrapping):} Sample $n$ indices uniformly with replacement $B$ times to create the ensemble.
    \item \textbf{Subbagging:} Sample $m$ indices, where $m < n$, uniformly without replacement $B$ times to create the ensemble.
\end{itemize}

\citet{breiman1996bagging} shows that bagging can be an effective way to achieve accuracy gains, claiming that the instability of the base predictor is the key that allows bagging to improve accuracy. \citet{buhlmann2002bagging} describe how bagging smooths hard decision boundaries and thus decreases variance and mean squared error. They introduce subbagging as a computationally cheaper alternative to bagging that achieves essentially identical performance. \citet{elisseeff2005} derive bounds on the stability of subbagged algorithms. More recently, \citet{soloff_bagging} have furthered this by providing additional guarantees on the stability of subbagged algorithms, though they remark that it does not necessarily imply that subbagging must improve upon the stability of the base algorithm. \citet{buja2006observations} show that bagging often, but not always, decreases variance while always increasing bias, at least in the U-statistic setting.

\citet{grandvalet2004bagging} explores the reasons why bagging works and why it occasionally fails. As an alternative to the threshold softening and variance reduction explanations, Grandvalet argues that bagging equalizes the influence of training samples. He refers to influence in the sense of the well-defined leverage in the simple regression setting --- that is, the sensitivity of a fitted value to a perturbation of the true label --- but also offers a generalized leverage statistic for non-linear models. According to this framework, it is the ``goodness/badness'' of highly influential samples that explains the effect of bagging on accuracy, be it an improvement or deterioration. In most situations, highly influential points are outliers to some degree and therefore bad leverage points, so the down-weighting of their influence by bagging typically reduces variance and improves performance as a downstream effect. Alternatively, when influential points are good leverage points, this same mechanism can increase variance and reduce performance. Here, an unstable predictor is one that has many highly influential points. This is why perturbing the data, and thus sometimes removing those points, leads to changes. This confirmed the hypothesis of \citet{breiman1996bagging} that bagging increases accuracy due to the instability of the base predictor, but clarified more thoroughly the mechanisms behind why this is the case. His results are supported by \citet{soloff_bagging}, in that ``the main stabilizing effect of bagging comes from the removal of high-leverage data points from a certain fraction of bags''.

\citet{grandvalet2006} expands on this idea for classification settings, where traditional measures of leverage do not apply. Hence, in the general case, he instead measures the influence of sample $i$ as 

\begin{equation}
\Delta(\hat{f}, \hat{f}^{\setminus i}) = \mathbb{P}_X\left(\hat{f}(X) \neq \hat{f}^{\setminus i}(X)\right)
\end{equation}

where $\hat{f}(X)$ and $\hat{f}^{\setminus i}(X)$ are the full and jackknife predictions, respectively, on a test set $X$. However, when a classifier returns prediction probabilities, a better measure of influence would be 

\begin{equation}
\Delta(\hat{p}, \hat{p}^{\setminus i}) = \mathbb{E}_X \left\|\hat{p}(X) - \hat{p}^{\setminus i}(X)\right\|
\end{equation}

where $\hat{p}(X)$ and $\hat{p}^{\setminus i}(X)$ are the full and jackknife prediction probability vectors, respectively.

Due to the essential equivalence of bagging and subbagging \citep{buja2006observations}, our results will largely be applicable to each approach, but for simplicity we will refer primarily to subbagging from now on. In the inflated argmax framework, subbagging is necessary to reduce the raw output instability, while the inflated argmax itself deals with the selection instability that remains due to test samples that sit near decision boundaries \citep{inflated_argmax}.

\section{The Rashomon Alternative}

The Rashomon set offers an alternative to resampling schemes as a way to generate multiple models from a single dataset. While bagging and subbagging average over numerous models of reduced quality due to resampling, the Rashomon set contains only high-quality models that are supported by the entire training set. This may help us avoid the situation where the subbagging process degrades accuracy to the point where any gains in stability are rendered meaningless.

The concept of the Rashomon effect was introduced by \citet{twocultures}, who observed that many distinct models can achieve near-equivalent predictive performance. \citet{Biecek_2024} illustrate this phenomenon with what they call a \textit{Rashomon Quartet}, which is a set of four models of different classes with identical predictive performance that nonetheless tell entirely different stories about the underlying relationships in the data. The idea and theory behind the Rashomon effect are outlined by \citet{parikh2026equallygoodmodelsanatomy}, who provides the following formal definition of the Rashomon set:

\begin{definition}[Rashomon Set]
\label{def:rashomon_set}
Let $\mathcal{F}$ denote a hypothesis class of models, $\mathcal{D} = \{(x_i, y_i)\}_{i=1}^n$ a 
dataset, and $L : \mathcal{F} \times \mathcal{D} \rightarrow \mathbb{R}^+$ a loss function. Let 
$f^* = \arg\min_{f \in \mathcal{F}} L(f, \mathcal{D})$ denote the empirical risk minimizer with 
optimal loss $L^* = L(f^*, \mathcal{D})$. The \textbf{Rashomon set} $\mathcal{R}(\epsilon, 
\mathcal{F}, \mathcal{D})$ for a given tolerance $\epsilon > 0$ is defined as:
\begin{equation}
    \mathcal{R}(\epsilon, \mathcal{F}, \mathcal{D}) = \{f \in \mathcal{F} : L(f, \mathcal{D}) 
    \leq L^* + \epsilon\},
\end{equation}
where $\epsilon$ is the Rashomon bound.
\end{definition}

In simpler terms, the Rashomon set is the collection of models that achieve nearly equivalent loss to that of the optimal model, with the loss for each model falling within $\epsilon$ of the optimal model. These models may be the optimal model under small perturbations, or they may never be the optimal model. The common trait that they all share is that their empirical risk differs negligibly from the optimal, for a small $\epsilon$, which means that accepting the optimal model as the ``truth'' implies that this negligible difference is meaningful. This is a sentiment that is difficult to justify.

The Rashomon ensemble takes a consensus of the predictions from each model in the Rashomon set to generate a final prediction. The idea is to average over the predictions from the Rashomon set as opposed to subbagged models, relying on the multiplicity to generate diverse predictions while using only the high-quality models within some Rashomon bound. \citet{realitrees} have shown that this approach can provide benefits for accuracy in active learning settings, so we extend this idea to stability.

If we are to leverage the Rashomon effect to create the ensemble, one potential issue is that there is no guarantee of diverse predictions from models within any given Rashomon set. Even if we have a Rashomon set of healthy size, it may be relatively homogeneous in nature and lack predictive multiplicity --- defined by \citet{pmlr-v119-marx20a} as ``the ability of a prediction problem to admit competing models with conflicting predictions''. We can think of a simple example where we have only two covariates with perfect collinearity, which would produce a Rashomon set that contains two models, one that relies on each variable. However, if we used these two models to make predictions, we would get the exact same result because both carry the same information. Hence, a Rashomon ensemble can ultimately be averaging over identical, or near-identical, prediction vectors, which clearly provides no benefit and adds computational load. As \citet{krogh} observe, combining the outputs of several predictors is only useful if they disagree on some inputs, and it is important for generalization that they disagree as much as possible. \citet{melville} build on this insight in the active learning space, showing that ensembles explicitly constructed for diversity can outperform standard bagging and boosting committees. Therefore, diversity is a critical property of an effective ensemble, and thus the most basic requirement is that the Rashomon set be diverse.

With resampling approaches like bagging and subbagging, the fact that we blind each model to some of the data, and thus degrade that iteration, is precisely what encourages diversity across the ensemble. While there is no guarantee of this within a Rashomon set, in many cases with a rich feature set, the Rashomon set will be relatively diverse in that it contains models that produce conflicting predictions. This diversity is what we want to leverage to garner more stability.

Naively ensembling the entire Rashomon set may result in the ensemble underutilizing the true diversity \citep{realitrees}. The Rashomon set may contain a plethora of trees that, while technically different, will have equivalent prediction vectors on the test set, and thus functionally they do not contribute anything new to predictive multiplicity. Hence, we need to prune the Rashomon set so that the ensemble does not get washed out by the most common prediction vector. We do this by removing any models that have duplicate prediction vectors and then adding a clustering algorithm to further condense the prediction set to truly unique predictive models. This helps us avoid the issue of insufficient diversity in the Rashomon set, which we know will doom the ensemble approach.

The other issue with using the Rashomon set has been the lack of computational efficiency. However, the SORTeD method from \citet{arslan2025sortedrashomonsetssparse} builds on previous decision tree Rashomon set approaches, like that of \citet{treefarms2022}, to the point that we can tractably run our experiments with full enumeration of exact Rashomon sets of decision trees. For the subbagging process, we need to generate enough subbagged models such that, for each perturbation, we have a sufficient number of subbagged models that do not include any left-out samples in the training data. For larger perturbations, this quickly becomes infeasible. The computational benefit of the Rashomon set ensemble is that we no longer need to do this. Rather, we must find the models in the Rashomon set for the full training dataset and for each perturbation resample. In this way, the Rashomon approach is actually simpler and potentially more efficient. However, this breaks down when we have many features, since the computational cost of the Rashomon set quickly increases with the number of features.

\section{Experimental Design}

The verbal autopsy data comes from the Population Health Metrics Research Consortium (PHMRC) and was collected in four countries: India, Mexico, Tanzania, and the Philippines \citep{phmrc2013}. The data contains survey responses and keyword counts, although our analysis is limited to the survey content. We condensed the cause-of-death categories down to five: cancers, cardiovascular, external injuries, infectious diseases, and other. The last category is a catch-all, but mostly contains noncommunicable and chronic diseases.

Therefore, we have the number of classes $L=5$ where $Y \in [L]$ represents the true labeled cause-of-death, determined by healthcare professionals. With Dar es Salaam, Tanzania designated as the validation site, the training set has a size of $n=6115$ and the test set contains $N=1726$ samples. For model configuration and hyperparameters, see Appendix~\autoref{app:models}.

We evaluate the performance of the following methods:
\begin{enumerate}
    \item The argmax of the base learning algorithm $\mathcal{A}$.
    \item The $\varepsilon$-inflated argmax of the base learning algorithm $\mathcal{A}$.
    \item The argmax of the subbagged algorithm $\widetilde{\mathcal{A}}_m$, with $B=500$ bags of size $m = n/2$.
    \item The $\varepsilon$-inflated argmax of the subbagged algorithm $\widetilde{\mathcal{A}}_m$, with $B=500$ bags of size $m=n/2$.
    \item The argmax of the Rashomon ensemble algorithm $\widetilde{\mathcal{A}}_{\mathcal{R}}$, with a majority vote from the Rashomon set with a maximum of 5,000 trees reduced to a set of distinct prediction vectors via clustering.
\end{enumerate}

Using the overlap definition of instability, we calculate the instability for each test point $j=1,\ldots,N$ across the $T$ perturbation models as

\begin{equation}
\delta_j := 1 - \frac{1}{T} \sum_{t=1}^{T} \frac{\left|s(\hat{p}(\tilde{X}_j)) \cap s(\hat{p}^{\prime}_t(\tilde{X}_j))\right|} {\min\left(\left|s(\hat{p}(\tilde{X}_j))\right|, \left|s(\hat{p}^{\prime}_t(\tilde{X}_j))\right|\right)} \, ,
\end{equation}

where $s$ is the selection rule applied to the prediction probabilities $\hat{p}$ --- from the model trained on the full dataset --- and prediction probabilities $\hat{p}^{\prime}_t$ --- from the model trained on the $t$-th perturbed dataset --- for the $j$-th test point $\tilde{X}_j$. For leave-one-out perturbations, we sample $T=500$ points $i_1,\dots,i_{500}$ uniformly without replacement from $[n]$ to create each leave-one-out instance. For leave-$k$-out perturbations, we sample $(n-k)$ indices uniformly without replacement from $[n]$, $T=200$ times. For bootstrap perturbations, we sample $n$ indices uniformly with replacement from $[n]$, $T=200$ times. The number of perturbed instances is reduced for the bootstrap and leave-$k$-out perturbations to compensate for the additional computational load. 

For leave-one-out perturbations, we are able to use a pooled approach, where we generate a large ``pool'' of bags and then, for each instance, select a subset of bags which do not contain the left-out point. This not only saves on computation, but it also reduces the sampling variation and better approximates the infinite bag instability effect. However, this is only possible with leave-one-out perturbations, where we can expect half the bags to remain under each perturbation, when $m = n/2$. It does not work with multi-point perturbations because the expected number of available bags becomes minuscule. Therefore, under multi-point perturbations, we must regenerate an entire set of bags for each instance, which is less efficient. 

With these instability measures, we take the mean and maximum instability across the test set, plus accuracy and set size measures to determine the full usefulness of each approach, as follows:

\begin{equation}
\beta_{\textnormal{mean-instability}} := \frac{1}{N} \sum_{j=1}^{N} \delta_j \, , \quad
\beta_{\textnormal{max-instability}} := \max_{j\in [N]} \delta_j \, .
\end{equation}

\begin{equation}
\beta_{\textnormal{accuracy}} := \frac{1}{N}\sum_{j=1}^N\mathbf{1}\left\{s(\hat{p}(\tilde{X}_j)) = \{{Y}_j\}\right\} \, , \quad
\beta_{\textnormal{coverage}} := \frac{1}{N}\sum_{j=1}^N\mathbf{1}\left\{\{{Y}_j\} \subseteq s(\hat{p}(\tilde{X}_j))\right\} \, ,
\end{equation} 

\begin{equation}
\beta_{\textnormal{size}} := \frac{1}{N}\sum_{j=1}^N \left|s(\hat{p}(\tilde{X}_j)) \right| \, .
\end{equation}

The accuracy, or more precisely singleton accuracy, is the proportion of the validation set for which the given model returns the true single class --- singleton set --- and thus is always inaccurate for any multi-class predictions. Coverage, on the other hand, is the proportion of the validation set for which the given model returns a prediction set that contains the true class. As the inflated argmax $\varepsilon$ parameter increases, we will trade off singleton accuracy for coverage while increasing the average size of the prediction sets.

In evaluating the Rashomon ensemble, we are much more severely restricted than with subbagging. Enumerating the Rashomon set for additional model classes is an ongoing research area, but for now a tree-based method is one of the few classes where exact enumeration is possible and the computation cost is reasonable. Subbagging, on the other hand, can be done with essentially any model class that is reasonably efficient.

With the SORTeD package \citep{arslan2025sortedrashomonsetssparse}, one could use the Rashomon multiplier or the maximum number of trees to restrict the size of the Rashomon set. However, since larger perturbations can alter the Rashomon effect, setting a maximum number of trees is more stable. For example, a fixed Rashomon multiplier may return a Rashomon set with thousands of trees for the full training data but then only return hundreds, or even just tens, of trees under some significant perturbation.

In our setting, the verbal autopsy dataset contains far too many features to make for an efficient example. Thus, to evaluate and compare the Rashomon ensemble approach, we instead turn to the Home Equity Line of Credit (HELOC) dataset from FICO and a synthetic dataset, the latter of which we can manipulate to effectively induce a strong Rashomon effect and predictive multiplicity \citep{fico2018heloc}. For the HELOC dataset, we have a training set of size $n=7844$ and a test set of size $N=2615$, from a random stratified split. For the synthetic dataset, $n=204$ and $N=560$, where the test set is augmented with ambiguous points (see Appendix~\autoref{app:synthetic}). With the full Rashomon set, we cluster the members' prediction vectors using k-means and then take a majority vote within each cluster to determine the prediction vector for each member of the final ensemble. Based on a sensitivity analysis, with results shown in \autoref{tab:cluster-sensitivity}, we chose the number of clusters to be 50 for the HELOC dataset and 10 for the synthetic dataset.

\section{Results}

\subsection{Stable and Unstable Classifiers}

\begin{figure}[!htbp]
    \centering
    \includegraphics[width=0.9\linewidth]{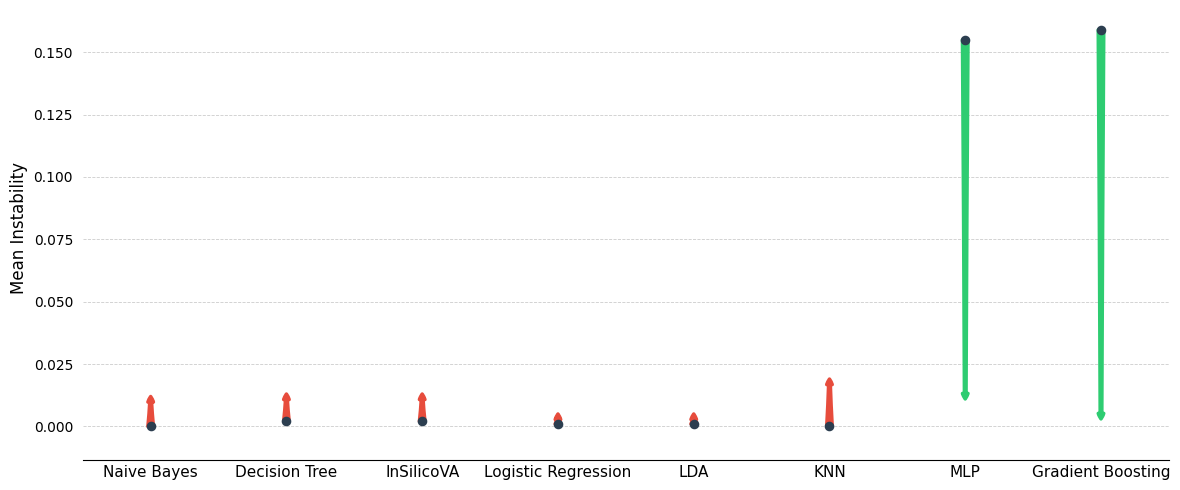}
    \caption{Comparing the effects of subbagging on mean instability ($\beta_{\textnormal{mean-instability}}$). The dots show the mean instability of the base algorithm ($\mathcal{A}$) and arrows point to the mean instability of the subbagged algorithm ($\widetilde{\mathcal{A}}_m$).}
    \label{fig:mean-comparison}
\end{figure}

In \autoref{fig:mean-comparison}, we can see the stark contrast in the effect of subbagging on stable and unstable classifiers. The stable classifiers have base mean instability near zero, and subbagging increases that slightly. The unstable classifiers start with relatively large mean instability, and subbagging provides a significant reduction, making them the perfect use case for subbagging. Illustrative examples are shown in the appendix in \autoref{tab:mlp} and \autoref{tab:gb}.

\begin{figure}[!htbp]
    \centering
    \includegraphics[width=0.9\linewidth]{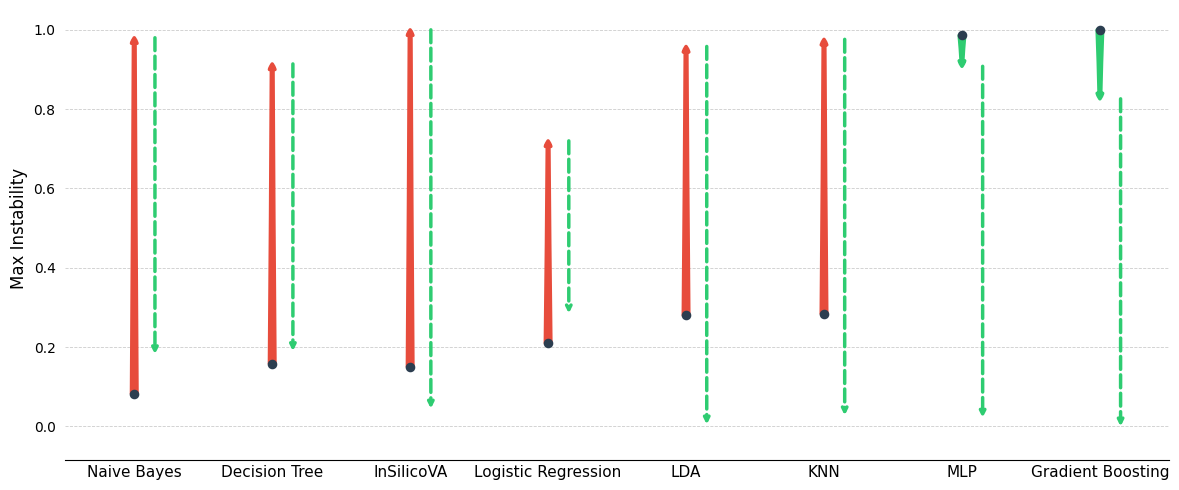}
    \caption{Comparing the effects of subbagging on maximum instability ($\beta_{\textnormal{max-instability}}$). The dots show the maximum instability of the base algorithm ($\mathcal{A}$), solid arrows point to the maximum instability of the subbagged algorithm ($\widetilde{\mathcal{A}}_m$), and dashed arrows point to the maximum instability of the $\varepsilon$-inflated subbagged algorithm ($\widetilde{\mathcal{A}}^{\varepsilon}_m$).}
    \label{fig:max-comparison}
\end{figure}

\autoref{fig:max-comparison} shows the effect of subbagging and the inflated argmax on maximum instability. With only subbagging, there is a significant increase in the maximum instability for the stable classifiers, whereas the unstable classifiers see minor reductions. The key point is that combining subbagging with the inflated argmax \textit{can} counteract the negative effects of subbagging a stable classifier and even bring the instability below that of the base algorithm. This will clearly not always be the case, since the inflated argmax may struggle with non-smooth classifiers and will not have much of an effect if the inflated argmax $\varepsilon$ parameter is not large enough. Furthermore, if we consider the Jaccard definition of stability, as seen in \autoref{fig:jaccard-comparison} of the appendix, then the effect of the inflated argmax is much less impressive. \autoref{tab:def-comparison} of the appendix offers a more detailed illustrative example of how each definition changes the effect of the inflated argmax, where the Jaccard definition is significantly more conservative.  

We found, as expected, that the inflated argmax alone often struggles in addressing instability issues. Subbagging by itself can significantly decrease the mean instability, though the maximum instability often remains rather high. However, combining subbagging and the inflated argmax most consistently reduces the instability to near zero.

\subsection{Increasing the Perturbation}
\label{more-perturbation}

The base stability is ultimately a relatively arbitrary measure, since leaving one sample out as a perturbation is an arbitrary choice. It could just as well be that we leave out more samples, thus causing larger perturbations. Then the instability of the base algorithm will inevitably be higher, yet nothing has fundamentally changed with our algorithm, so should the effect of subbagging be any different for the supposedly stable classifiers?

To investigate this question, we increase the perturbation level to see how that changes the effect of subbagging. Instead of a minor leave-one-out perturbation, we perform leave-$k$-out perturbations, with $k$ increasing. We also perform bootstrap resampling, which provides both a higher magnitude of perturbation and a different perturbation regime. When using either of these more significant perturbation schemes, the base instability of each algorithm will inevitably be \textit{nominally} higher than with leave-one-out since the magnitude of the perturbation is higher. The interesting part is what happens with the effect of subbagging.

For leave-$k$-out perturbations, as we increase $k$, the effect of subbagging can take divergent paths for classifiers that are jackknife stable. With the Linear Discriminant Analysis (LDA) classifier, the stability curves converge and the effect of subbagging becomes negligible. With the decision tree classifier, on the other hand, subbagging goes from being quite detrimental under leave-one-out perturbations to noticeably beneficial under leave-$k$-out perturbations, for higher values of $k$. This effect can be seen in \autoref{fig:leave-k-out-grid} in the appendix.

The same effect can be seen with bootstrap perturbations, as shown in \autoref{fig:bs-instability} below (LDA and decision tree) and \autoref{fig:app-bs-instability} in the appendix (naive Bayes and regularized decision tree), all of which are jackknife stable. Each of these classifiers saw subbagging increase the instability when considering leave-one-out perturbations. For the non-tree models, when we increase the perturbation level, the instability of each approach converges, at least approximately, where the instability is essentially equivalent across each procedure. In these cases, our chosen perturbation level changed the \textit{magnitude} of the effect of subbagging, but did not change the direction of that effect.

\begin{figure}[!htbp]
    \centering
    \begin{subfigure}{0.48\textwidth}
        \centering
        \includegraphics[width=\textwidth]{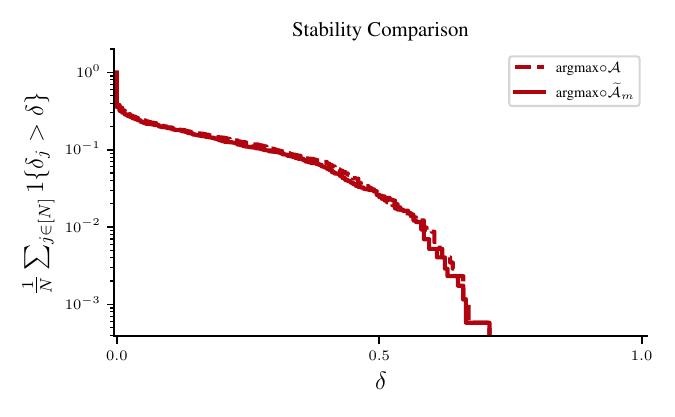}
        \caption{Linear Discriminant Analysis (LDA)}
        \label{fig:bs-left}
    \end{subfigure}
    \hfill
    \begin{subfigure}{0.48\textwidth}
        \centering
        \includegraphics[width=\textwidth]{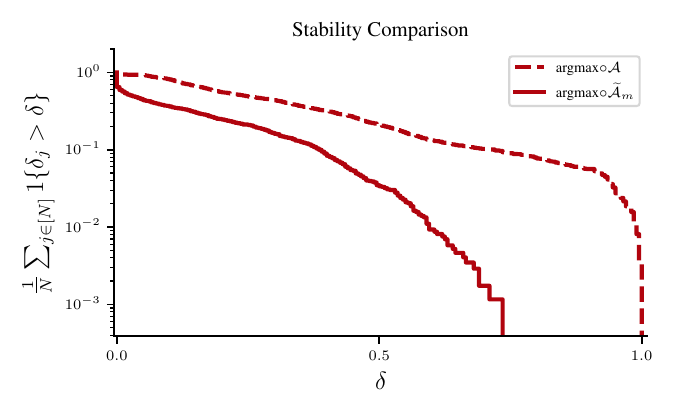}
        \caption{Decision Tree}
        \label{fig:bs-right}
    \end{subfigure}
    \caption{Instability plots for jackknife-stable classifiers with bootstrap perturbations and $B=500$. The curves show the proportion of samples in the test set with instability greater than $\delta$ for unbagged (dashed) and subbagged (solid) algorithms.}
    \label{fig:bs-instability}
\end{figure}

However, with these tree classifiers, this convergence is not observed. Instead, the directional effect of subbagging reverses from being detrimental to becoming an improvement. It is unclear whether this effect is specific to trees. Regardless, we see that the chosen perturbation level does matter in determining the effect of subbagging. In some cases, larger perturbation levels render subbagging meaningless, while in others larger perturbation levels may allow subbagging to actually improve stability. For our decision tree classifier, subbagging increased the mean leave-one-out instability from 0.002 to 0.013 but decreased the mean bootstrap instability from 0.317 to 0.125. In scenarios like this, we may want to sacrifice a small increase in jackknife instability if it comes with a significant enough reduction in bootstrap instability, especially if the inflated argmax can further reduce both types of instability to near zero.

\subsection{Subbagging Fraction and Inflated Argmax Sensitivity}

Choosing the correct subbagging fraction and $\varepsilon$-inflated argmax parameter is not governed by a simple formula. In practice, the best approach is likely to be trial-and-error based. First, depending on the class of model and the dataset, computation can be quite expensive, especially as $m$ increases. This cost makes a thorough exploration relatively infeasible. There is also no easy way to gauge the correct $\varepsilon$ parameter without first testing values, since we cannot directly target a certain set size, for example. Finally, different scenarios are likely to present different goals. One setting may favor coverage, while another favors singleton accuracy. Ultimately, different practitioners may want to target different metrics.

\begin{figure}[!htbp]
    \centering
    \begin{subfigure}{0.48\textwidth}
        \centering
        \includegraphics[width=\textwidth]{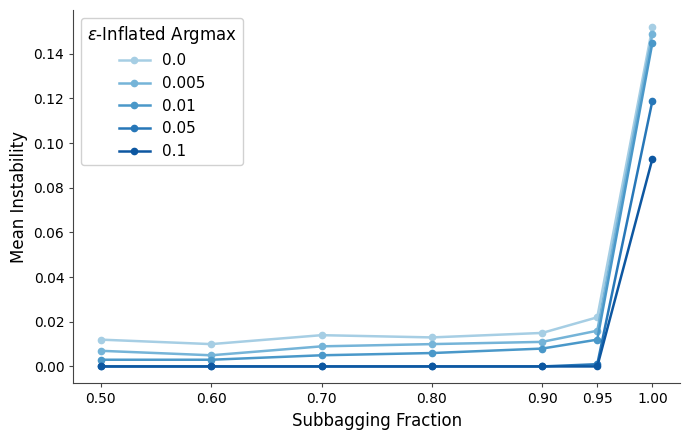}
        \caption{}
        \label{fig:ins-left}
    \end{subfigure}
    \hfill
    \begin{subfigure}{0.48\textwidth}
        \centering
        \includegraphics[width=\textwidth]{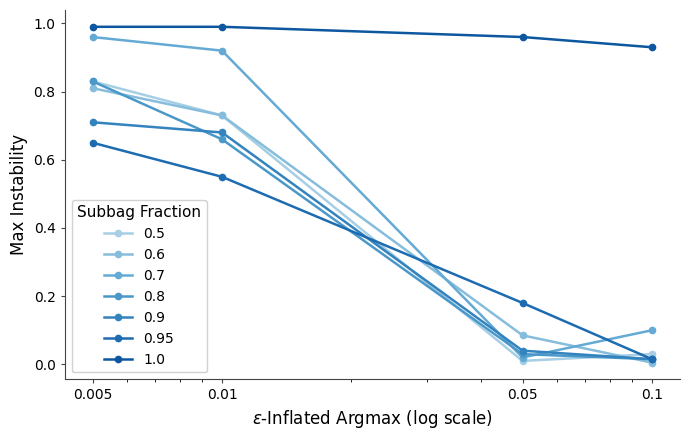}
        \caption{}
        \label{fig:ins-right}
    \end{subfigure}
    \caption{Subbagging fraction (${m}/{n}$) and $\varepsilon$-inflated argmax sensitivity analysis. The results illustrate the parameter ranges over which the subbagging fraction and the $\varepsilon$-inflated argmax most significantly reduce instability.}
    \label{fig:sensitivity-plots}
\end{figure}

In our sensitivity analysis, the first observation is that subbagging does not necessarily require significant subsampling to induce the stability benefits. \autoref{fig:ins-left} shows that for the MLP classifier, the instability is about as low with 95\% of the data sampled for each bag as it is with 50\% of the data. This gives significant leeway in choosing the subbagging fraction, which is especially useful in settings where the fraction has a noticeable effect on accuracy. See \autoref{fig:sensitivity-plots-extra} in the appendix for the sensitivity with respect to the other metrics.

Furthermore, \autoref{fig:ins-right} shows that while subbagging is often the significant driving force when it comes to \textit{mean} instability, the $\varepsilon$ parameter of the inflated argmax is more influential when it comes to \textit{maximum} instability. The maximum instability deals with the worst-case, or most ambiguous, sample in the test set. These points are likely to be near the boundary between classes. Although subbagging alone struggles to effectively handle that, the inflated argmax is built specifically to solve this issue by including all classes close enough to the boundary. We can also clearly see that the inflated argmax does not work well without any resampling.

\subsection{Stability of the Rashomon Ensemble}

Although we are restricted to a single model class for the Rashomon ensemble and thus cannot make comparisons across a broad range of stable and unstable classifiers, we still find promising, if not ultimately mixed, results. 

In our initial experiments, leave-one-out perturbations were simply not significant enough to noticeably alter the model or the Rashomon set, given the cost complexity imposed to keep the Rashomon computation reasonable (see \autoref{fig:rashomon-plots-loo} in the appendix). Since subbagging is ineffective in this case, it is unreasonable to expect the Rashomon ensemble to perform any better. Yet, as we saw previously, specifically with tree-based methods, even if the classifier is jackknife stable, it can still be bootstrap unstable and the subbagging ensemble approach can help mitigate that. Hence, we evaluate the performance of the Rashomon ensemble method under leave-$k$-out and bootstrap perturbations.

\begin{figure}[!htbp]
    \centering
    \begin{subfigure}{0.49\textwidth}
        \centering
        \includegraphics[width=\textwidth]{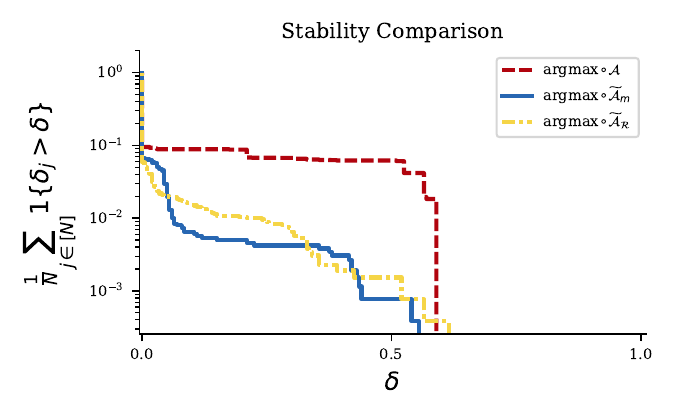}
        \caption{HELOC Dataset, $k$ = 1000, $n$ = 7844}
        \label{fig:ins-rashomon-heloc-k}
    \end{subfigure}
    \hfill
    \begin{subfigure}{0.49\textwidth}
        \centering
        \includegraphics[width=\textwidth]{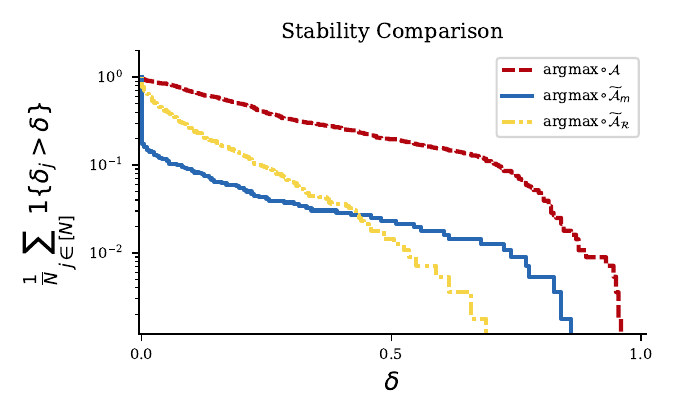}
        \caption{Synthetic Dataset, $k$ = 25, $n$ = 204}
        \label{fig:ins-rashomon-synth-k}
    \end{subfigure}
    \caption{Instability plots from leave-$k$-out perturbations with the Rashomon ensembles based on 5,000 trees. The curves show the proportion of samples in the test set with instability greater than $\delta$.}
    \label{fig:rashomon-plots-k-out}
\end{figure}

\autoref{fig:rashomon-plots-k-out} shows the results from the HELOC dataset (left) and the synthetic dataset (right). In the HELOC example, we see that the Rashomon ensemble can provide clear benefits in terms of stability. Subbagging and the Rashomon ensemble achieve near-identical reductions in instability, with both dropping mean instability from 0.041 to 0.005. On the full training set, the Rashomon set contained 5,497 trees (the maximum is not strict, likely accepting ties), of which 347 produced unique prediction vectors. The results with bootstrap perturbations are similar and can be seen in \autoref{fig:rashomon-plots-bs} in the appendix.

Although the Rashomon ensemble did match subbagging in the HELOC case, we could not find an example where it significantly outperformed subbagging. In our experiments with other datasets, it was more often the case that both approaches performed similarly while not providing much instability reduction, though this may be more of a reflection on the method underlying SORTeD.

The synthetic dataset was designed to induce a strong Rashomon effect by pruning training points near the decision boundary — the samples most informative about boundary placement — leaving the boundary region sparse and allowing many distinct models to achieve near-identical empirical loss. If there existed different mechanisms driving instability, then this dataset should target a multiplicity mechanism, if it exists and is separate from the random sampling noise that subbagging addresses. On the full training set, the Rashomon set contained 6,573 trees, of which 1,373 produced unique prediction vectors. As we see in \autoref{fig:ins-rashomon-synth-k}, the Rashomon ensemble clearly reduces instability. It is more effective at reducing the maximum instability than subbagging is --- dropping maximum instability from 0.96 to 0.69 for the Rashomon ensemble versus 0.86 for subbagging. However, it fails to beat subbagging in terms of mean instability reduction, from 0.275 down to 0.080 for the Rashomon ensemble versus 0.033 for subbagging. This provides little evidence of separate mechanisms, since even in the setting designed for the Rashomon ensemble, it is not the dominant approach.

When comparing fits on the same dataset with different parameters, we found that increasing the predictive multiplicity, as measured by average pairwise disagreement within the Rashomon set or within the pruned ensemble, does not necessarily translate to a greater reduction in instability (see \autoref{tab:disagree1} and \autoref{tab:disagree2}). To be fair, predictive multiplicity is hard to condense into a single metric, so pairwise disagreement is one operationalization of it. We know that predictive multiplicity is necessary, to some degree, but there is no consistent and direct relationship between it and instability reduction that we have found.

These examples serve to show that it is \textit{possible} for the Rashomon ensemble to be effective under favorable conditions, not that this is necessarily typical or that it is necessarily better than alternative techniques. It can reduce instability and match or, in some respects, surpass the subbagging ensemble in performance, but it is not a valid substitute for subbagging in general since it does not dominate in any observed scenario. As additional Rashomon enumeration methods are developed, we believe that these results, specific to the class of sparse decision trees, will generalize to other model classes as well and could provide a clearer picture of exactly how these two ensemble methods compare.

\section{Discussion}

\subsection{When Ensembling Succeeds}

The stability of a learning algorithm under leave-one-out perturbations is directly related to the influence of each of the left-out samples, under the jackknife definition of influence, since both measure the sensitivity of predictions to the leave-one-out perturbation. Thus, for any technique to improve stability, it must in aggregate reduce the influence of samples in the training set. A classifier that is jackknife unstable has many high-influence points, so, under the \citet{grandvalet2004bagging} interpretation, the down-weighting of these high-influence points is the dominant effect and leads to a reduction in overall influence, which directly translates to jackknife stability. Subbagging excludes a certain proportion of data points from each bag, giving those points no influence. Thus, when these high-influence points are left out as a perturbation, the predictions from many of the bags do not change, thereby providing stability. 

The Rashomon ensemble has no resampling scheme, but there is still a clear influence equalization effect, which we see in \autoref{fig:infl-rashomon-2} of the appendix. Referring to \autoref{def:rashomon_set}, the hypothesis class of models $\mathcal{F}$ is fixed and the dataset $\mathcal{D}$ enters only through the loss function $L$ to determine membership in the Rashomon set $\mathcal{R}(\epsilon, \mathcal{F}, \mathcal{D})$. By defining $\mathcal{F}$ over the feature space $\mathcal{X}$, the hypothesis space from which ensemble membership can be chosen is invariant to data perturbations such as leave-one-out. The implication is that, unlike stochastic resampling schemes, data perturbations are not accompanied by refitting a model, but rather apply a new filter or aperture that shifts across a pre-existing hypothesis space. Any given model $f \in \mathcal{F}$ has no sensitivity to perturbations; it remains static.

\citet{treefarms2022} show that the Rashomon set at $\epsilon'$-tolerance after removing a group of instances is contained within the full-dataset Rashomon set at a larger $\epsilon$-tolerance. Hence, under a data perturbation such as leave-one-out, denoted $\mathcal{D}^{\setminus i}$, we have $\mathcal{R}(\epsilon', \mathcal{F}, \mathcal{D}^{\setminus i}) \subseteq \mathcal{R}(\epsilon, \mathcal{F}, \mathcal{D})$. Since Rashomon sets are nested and $\epsilon' \leq \epsilon$, we also have $\mathcal{R}(\epsilon', \mathcal{F}, \mathcal{D}^{\setminus i}) \subseteq \mathcal{R}(\epsilon, \mathcal{F}, \mathcal{D}^{\setminus i})$. Together, these imply that $\mathcal{R}(\epsilon, \mathcal{F}, \mathcal{D}) \cap \mathcal{R}(\epsilon, \mathcal{F}, \mathcal{D}^{\setminus i})$ is non-empty. For all models $f$ that are contained in the perturbed and unperturbed Rashomon sets at the $\epsilon$-tolerance, such that $f \in \mathcal{R}(\epsilon, \mathcal{F}, \mathcal{D}) \cap \mathcal{R}(\epsilon, \mathcal{F}, \mathcal{D}^{\setminus i})$, these committee members will contribute the same information regardless of the perturbation. Therefore, the jackknife influence of the removed point will be zero for these persistent committee members, which provides an effect akin to the exclusion of points that occurs in bagging. 

In the Rashomon ensemble, for a fixed $\epsilon$, a data perturbation shifts the membership boundary relative to the fixed elements of $\mathcal{F}$ by altering the loss $L(f,\mathcal{D})$ for every $f \in \mathcal{F}$. So, the influence of a removed point is only manifest in the fluid margin of the ensemble --- the models that enter or exit the Rashomon set threshold due to the perturbation. A subset of the original Rashomon set will remain and thus the stable intersection dilutes the influence, mimicking the down-weighting effect seen in resampling schemes. The theorem from \citet{treefarms2022} applies directly to leave-$k$-out perturbations, though not to bootstrap perturbations since it deals with fixed group removal. Still, we conjecture that it remains relevant, perhaps to a lesser degree, since it follows the construction of the Rashomon set under data perturbations. This is supported by the rough equivalence of outcomes with leave-$k$-out and bootstrap perturbations.

Unlike subbagging, this effect is far from guaranteed and requires not insignificant stable intersections. In our experiments, we found that a strong Rashomon effect, with sufficient predictive multiplicity, is necessary for a meaningful reduction in instability. However, predictive multiplicity of the Rashomon set alone is not enough. Models must also vary in which points drive their membership so that a stable intersection can persist under perturbation. We found no evidence that the Rashomon ensemble can significantly outperform subbagging, but we are restricted to just a single tree-based class of models. In the future, a fairer comparison can be made once we are able to efficiently enumerate Rashomon sets across many different model types, which may reveal settings in which the Rashomon ensemble is far more effective than we have shown.

\subsection{When Ensembling Fails}

We argue, as the results of our experiments have shown, that the issue with both ensemble approaches is that they are detrimental when the base algorithm --- without ensembling --- is already stable. Subbagging is known to increase bias and can sometimes increase variance \citep{buja2006observations}. The bias issue can affect accuracy, but does not explain the negative effects on stability since a biased subbagged model will induce similarly biased jackknife models, so the bias effect should largely cancel out in the instability comparison.

Increased variance could serve as an explanation, but the issue is that the effects of subbagging on accuracy and stability do not necessarily move in the same direction. \autoref{tab:tree-lc} of the appendix shows that even when subbagging significantly increases accuracy, it can still negatively impact stability --- with a minor effect on mean instability but a large effect on maximum instability. As \citet{breiman1996bagging}, \citet{buhlmann2002bagging}, and \citet{grandvalet2004bagging} all agree, when subbagging aids predictive performance, it is associated with a reduction in variance. Hence, an increase in variance is also not sufficient to explain why the instability increased.

Regardless of the state of the base algorithm, subbagging should theoretically always equalize influence, and if the influence is already relatively equal, then there should be no effect, as argued by \citet{grandvalet2004bagging}. However, even if they are often associated, \textit{equalizing} influence does not necessarily equate to \textit{reducing} influence. \autoref{fig:influence-plots} of the appendix shows that even for jackknife-stable classifiers, there is still an influence equalization effect with subbagging, even though the instability increases and thus the overall influence increases as well.

To explain this phenomenon, we observe that the increase in instability due to subbagging persists even without any perturbation to the data. As \autoref{fig:instability-plots-no-perturb} shows, for a jackknife-stable classifier, the instability effect associated with subbagging under leave-one-out perturbations remains approximately equal even with no perturbation at all. This suggests that the practice of selecting a finite number of bags injects some variance that is not present in the standard algorithm. Furthermore, this noise floor is not an artifact of any particular implementation. Whether bags are drawn from a shared pool or fully regenerated for each instance, each comparison necessarily uses a different finite draw of bags. When a perturbation is present, we cannot hold the selected bags fixed while also excluding, for each instance, the bags containing the left-out point(s). So, for a fair comparison, we proceed with the same bag selection procedure, as would be necessitated under a perturbation, but simply do not perform the perturbation. This produces two instability curves that are nearly indistinguishable.

This noise arises because each individual bag is already a much larger perturbation of the training data than a leave-one-out perturbation. Subbagging with fraction $m/n$ removes $(n-m)$ points from each bag, while jackknife stability only characterizes sensitivity to removing a single point. Clearly, stability under leave-one-out perturbations does not imply stability under the removal of a much larger fraction of the data, so even a jackknife-stable classifier can disagree substantially, bag to bag, from the fit on the full dataset. With only finitely many bags, these disagreements do not fully average out, producing instability even without any perturbation applied. In the case of a jackknife-unstable base algorithm, this effect is negligible compared to the influence equalization and reduction. With little or no reduction needed, this destabilizing effect now dominates.

To be clear, this is not the classical predictive variance that governs the bias-variance tradeoff, but Monte Carlo noise from using a finite number of bags $B$. According to \cite{inflated_argmax}, 

\begin{theorem}
For any base learning algorithm $A$ returning outputs in $\Delta_{L-1}$, the bagged algorithm $\widetilde{A}_{m}$ (computed with a finite number of bags, $B$) has tail stability $(\varepsilon, \delta)$ for any $\varepsilon, \delta > 0$ satisfying
\[
\delta = \varepsilon^{-2} \cdot (1 - 1/L) \left( \frac{1}{n-1} \cdot \frac{p_{n,m}}{1-p_{n,m}} + \frac{16e^2}{B} \right), \, \text{ where } p_{n,m} = \frac{m}{n}.
\]
\end{theorem}

With finite bags, $B$, we have an extra term that only goes to zero as $B \to \infty$. However, while the theorem implies that the tail stability decreases monotonically as $B \to \infty$, this describes an upper bound. The empirical instability does not behave monotonically, and thus the asymptotic behavior can be difficult to achieve in practice. In \autoref{fig:instability-plots-no-perturb}, the Monte Carlo noise persists even with $B=2500$ bags and the maximum instability remains high at 0.985, which is even higher than the maximum instability of 0.980 with $B=500$ bags. So, a 5-fold increase in $B$ produced no discernible reduction in the worst-case instability, illustrating how slowly the noise floor shrinks in practice. Whether this noise changes a sample's predicted class depends on its margin, being the gap between the top class's predicted probability and that of the runner-up. For the most ambiguous test points, this margin can be close to zero, such that even a small amount of noise is enough to flip the predicted class. Driving the noise below this margin requires increasing $B$, but because the margin for the worst-case point can be arbitrarily small, the number of bags needed to guarantee stability for that point can be correspondingly large, and in some datasets, impractically so. Even if subbagging has little or no detrimental effect asymptotically, with a finite and feasible number of bags it can remain a real practical concern.

A destabilizing mechanism is clearly present with the Rashomon ensemble, as observed in \autoref{fig:rashomon-plots-loo} of the appendix. However, as \autoref{fig:rashomon-plots-no-perturb} shows, it is not just Monte Carlo noise. The unperturbed procedure has zero instability by construction, since there is no stochastic step. Given that deterministic structure, any instability observed under perturbation must be attributable to shifts in Rashomon set membership induced by the perturbation, with no finite-sample noise component. When the algorithm for the optimal model is jackknife stable, the loss of the optimal model, $L^*$, will not change much under leave-one-out perturbations. However, knowing that the optimal loss $L^*$ is relatively insensitive to perturbation does not guarantee the same invariance for $L(f, \mathcal{D})$ across all $f \in \mathcal{F}$. For a fixed $f \in \mathcal{F}$, removing point $i$ shifts $L(f, \mathcal{D})$ by an amount governed by how $f$'s own loss at point $i$ compares to $f$'s average loss overall. So, for models $f \in \mathcal{F}$ that sit near the Rashomon membership boundary, those that fit point $i$ exceptionally well or exceptionally poorly are those whose membership is most likely to flip. Even for a small change in the loss, if the model crosses the boundary, then the ensemble gains or loses an entire prediction vector. Because the baseline instability is already near zero in this case, even a single such entry or exit is no longer negligible. The churn that would be dwarfed by a high baseline for an unstable base algorithm instead dominates the ensemble average for a stable one, thus inducing the destabilizing effect.

These explanations fit with the results of \autoref{more-perturbation}, where we saw that, for jackknife-stable classifiers, the stability of a subbagged algorithm can converge to that of the base algorithm as the perturbation level increases, if not reverse the directional effect entirely. This is consistent with the interpretation that when the perturbation level increases, the nominal instability of the base algorithm increases away from zero, and thus the noise floor ceases to dominate. However, since the base algorithm is still inherently jackknife stable, no single point has outsized influence, and thus the benefit may be limited. The higher instability is induced by a more significant perturbation, where high-influence clusters of points are removed. As \citet{grandvalet2004bagging} points out, bagging can be less effective at down-weighting clusters of leverage points because the probability of excluding the entire cluster from a bag becomes exceedingly small with larger clusters. Hence, bagging and subbagging may struggle to even affect a classifier that is jackknife stable yet bootstrap unstable because of the nature of the resampling processes. This is likely why, in some cases, we see subbagging have no effect at these higher perturbation levels. 

This is where the Rashomon ensemble could have an advantage because it has no such drawback and may be able to effectively down-weight clusters of leverage points by emphasizing different features. This warrants further investigation, but again we are currently restricted to tree-based methods, which unfortunately do not represent the setting where the effect of subbagging disappears at higher perturbation levels. As such, we require a new enumeration method that relies on a different model class.

In many scenarios, achieving jackknife stability is sufficient, but this is certainly not always the case, especially with larger sample sizes. In settings where we worry about what happens under more significant perturbations, the effect of subbagging on jackknife-stable classifiers can either be completely negated or actually become beneficial. When subbagging is rendered ineffective at combating significant perturbations, it is not necessarily clear that the Rashomon ensemble would suffer the same effect. On the other hand, when subbagging is detrimental to jackknife stability but beneficial to bootstrap stability, a tradeoff must be made regardless of the ensemble method.

\subsection{Recovering Stability with the Inflated Argmax}

In our experiments, output stability mostly mirrored selection stability, though it is important to remember that this is not always the case. Subbagging has a direct effect on \textit{output} stability and thus an indirect effect on selection stability. The inflated argmax rule deals specifically with selection stability, as it has no effect on the raw output.

The inflated argmax approach can actually remedy the negative effect that ensembling may have on stability. We saw that even when subbagging increased the selection instability, adding the inflated argmax on top of it could reduce that instability down to essentially zero, even sending the maximum instability below that of the original base algorithm in some cases. The added instability associated with subbagging is relatively minor and predominantly due to Monte Carlo noise that is unlikely to generate large shifts. Therefore, most of the added instability should be from points that are still near decision boundaries. Hence, the inflated argmax can help, which allows us to utilize subbagging on any classifier, be it unstable or stable, without sacrificing much stability. Granted, this may not be true with the Jaccard definition of instability, which is where it falls to the user to determine the definition of instability that fits the goals of the experiment.

\section{Conclusion}

Instability is a direct product of the high influence of individual data points in the training set, such that small jackknife perturbations can lead to significant changes when these points are removed. Our results are consistent with ensemble approaches --- from bagging and subbagging to the Rashomon ensemble --- equalizing the influence of data points. The Rashomon ensemble does not have the same diversity guarantee as subbagging and thus works only in settings with predictive multiplicity and diversity of influence. Currently, the Rashomon ensemble is not a reasonable substitute for subbagging because not only does it fail to outperform subbagging in general, but it needs a specific Rashomon framework and cannot be applied to any model class like subbagging can. That being said, there is potential efficacy in the Rashomon ensemble and we hope that it can improve as the capacity to enumerate Rashomon sets for all algorithms improves. 

When a base algorithm is jackknife unstable, these ensemble techniques provide a stabilizing effect, yet they provide a destabilizing effect when the base algorithm is already stable. This is likely due to the noise introduced by the ensemble itself, be it finite-bag Monte Carlo variation for subbagging or churn in Rashomon set membership. While this noise is negligible relative to high baseline instability, it becomes the dominant effect when the baseline is already low. For selection stability specifically, the inflated argmax smooths hard decision boundaries and can even counteract the destabilizing forces associated with ensembling stable classifiers, potentially making ensembling applicable and beneficial in more settings.

\subsubsection*{Acknowledgments}

We would like to thank Hayden McTavish and Cynthia Rudin for valuable discussions and suggestions. This project was supported by grants from the National Science Foundation under Award No. SES-2215369 and was sponsored by the Department of the Navy, Office of Naval Research, under Award No. N00014-25-1-2270. Any opinions, findings, and conclusions or recommendations expressed in this material are those of the author(s) and do not necessarily reflect the views of the National Science Foundation or the Office of Naval Research. We also thank the National Bureau of Economic Research for critical administrative support.

\bibliography{main}
\bibliographystyle{tmlr}

\appendix
\section{Appendix}

\subsection{Figures}
\label{app:figures}

\subsubsection{Subbagging}
\label{app:subbagging-figures}

\begin{figure}[H]
    \centering
    \includegraphics[width=0.9\linewidth]{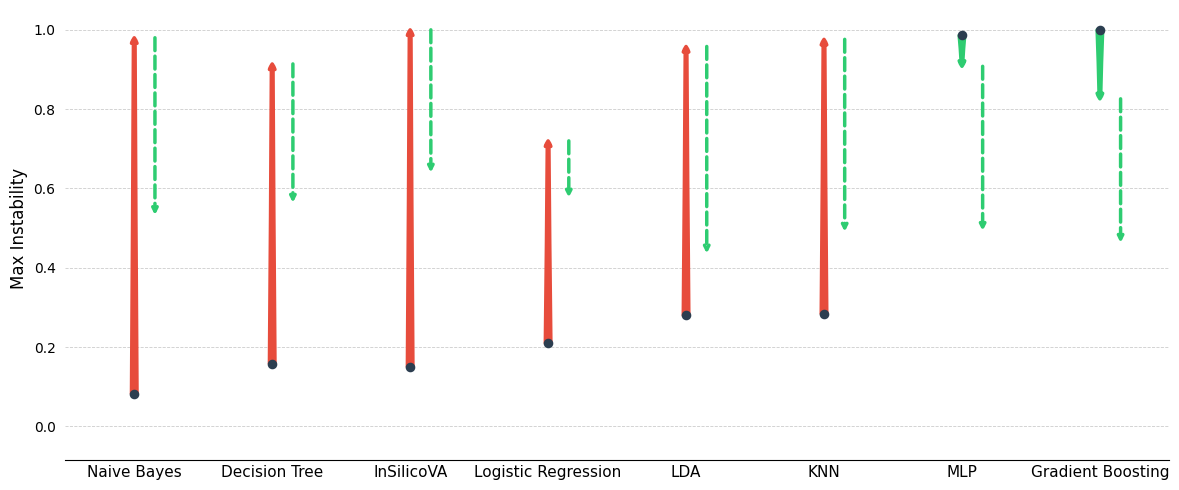}
    \caption{Comparing the effects of subbagging on maximum instability ($\beta_{\textnormal{max-instability}}$) with Jaccard instability. The dots show the maximum instability of the base algorithm ($\mathcal{A}$), solid arrows point to the maximum instability of the subbagged algorithm ($\widetilde{\mathcal{A}}_m$), and dashed arrows point to the maximum instability of the $\varepsilon$-inflated subbagged algorithm ($\widetilde{\mathcal{A}}^{\varepsilon}_m$). Under the Jaccard definition of instability, the inflated argmax does not provide as significant a reduction in instability as it does with the disjoint or overlap definitions.}
    \label{fig:jaccard-comparison}
\end{figure}

\begin{figure}[H]
    \centering
    \begin{subfigure}{0.48\textwidth}
        \centering
        \includegraphics[width=\textwidth]{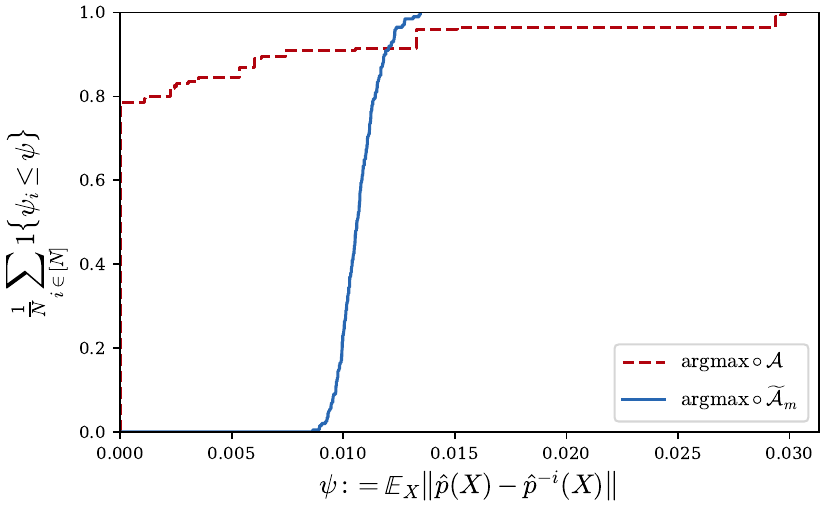}
        \caption{InSilicoVA}
        \label{fig:influence_left}
    \end{subfigure}
    \hfill
    \begin{subfigure}{0.48\textwidth}
        \centering
        \includegraphics[width=\textwidth]{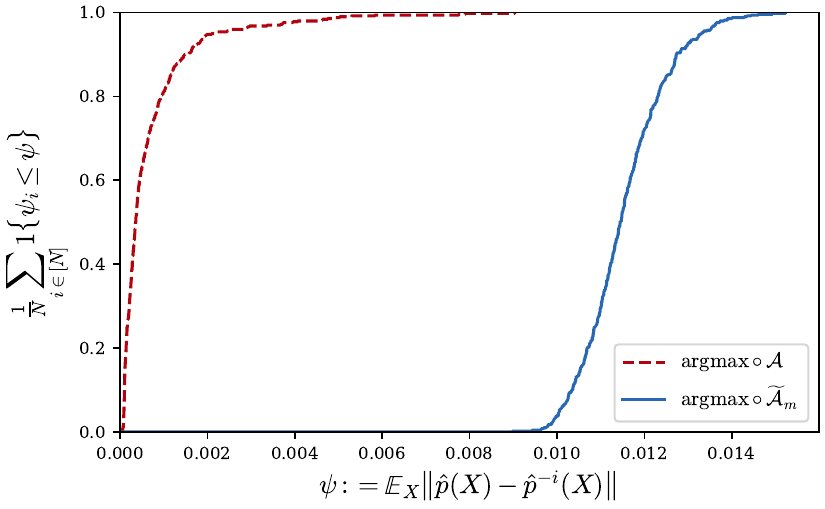}
        \caption{Naive Bayes}
        \label{fig:influence_right}
    \end{subfigure}
    \caption{Empirical cumulative distribution function of influence as defined by $\mathbb{E}_X \|\hat{p}(X) - \hat{p}^{\setminus i}(X)\|$ for unbagged (dashed) and subbagged (solid) algorithms. Influence equalization does not imply instability reduction since these are examples where the instability increases with subbagging.}
    \label{fig:influence-plots}
\end{figure}

\begin{figure}[H]
    \centering
    \captionsetup[subfigure]{font=small, skip=0.1pt}
    \begin{subfigure}{0.46\textwidth}
        \centering
        \includegraphics[width=\textwidth]{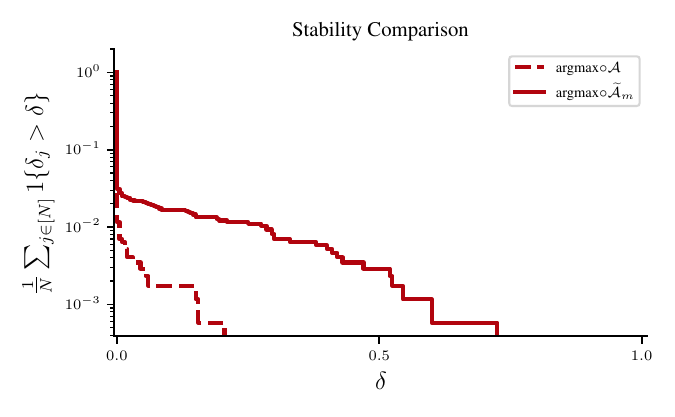}
        \caption{LDA, $k=1$}
        \label{fig:k-grid-lda-1}
    \end{subfigure}
    \hfill
    \begin{subfigure}{0.46\textwidth}
        \centering
        \includegraphics[width=\textwidth]{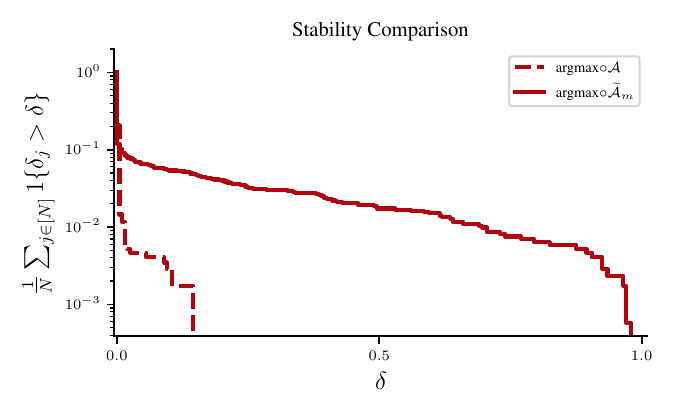}
        \caption{Decision Tree, $k=1$}
        \label{fig:k-grid-dt-1}
    \end{subfigure}

    \vspace{0.2em}
    \begin{subfigure}{0.46\textwidth}
        \centering
        \includegraphics[width=\textwidth]{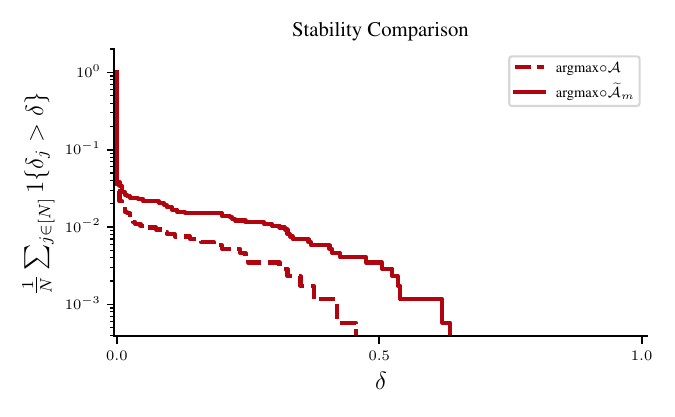}
        \caption{LDA, $k=10$}
        \label{fig:k-grid-lda-10}
    \end{subfigure}
    \hfill
    \begin{subfigure}{0.46\textwidth}
        \centering
        \includegraphics[width=\textwidth]{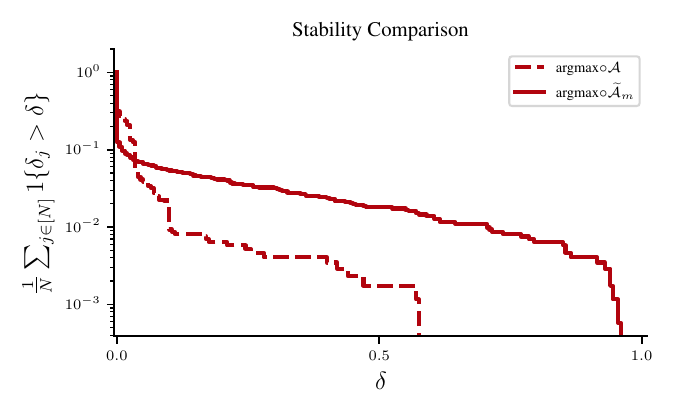}
        \caption{Decision Tree, $k=10$}
        \label{fig:k-grid-dt-10}
    \end{subfigure}

    \vspace{0.2em}
    \begin{subfigure}{0.46\textwidth}
        \centering
        \includegraphics[width=\textwidth]{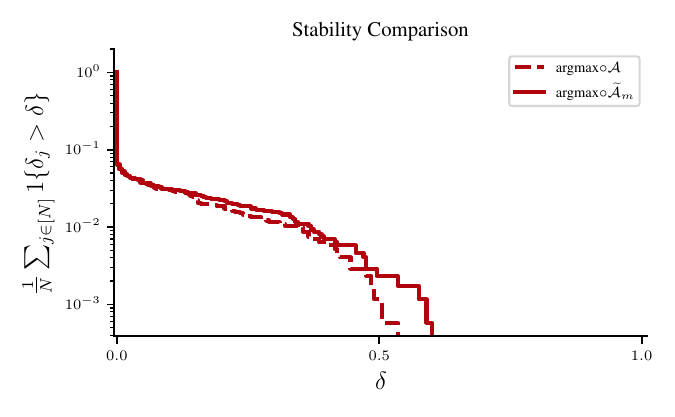}
        \caption{LDA, $k=100$}
        \label{fig:k-grid-lda-100}
    \end{subfigure}
    \hfill
    \begin{subfigure}{0.46\textwidth}
        \centering
        \includegraphics[width=\textwidth]{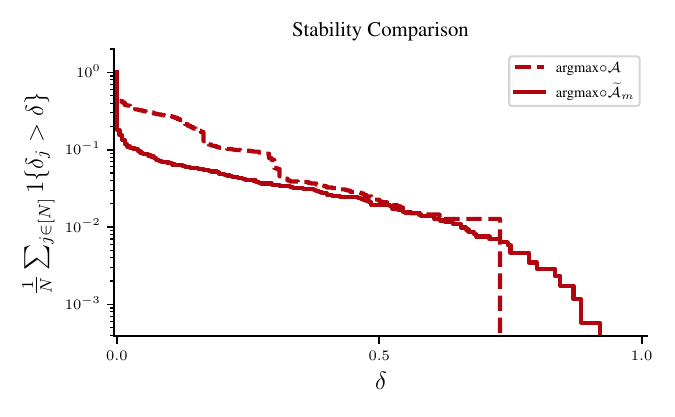}
        \caption{Decision Tree, $k=100$}
        \label{fig:k-grid-dt-100}
    \end{subfigure}

    \vspace{0.2em}
    \begin{subfigure}{0.46\textwidth}
        \centering
        \includegraphics[width=\textwidth]{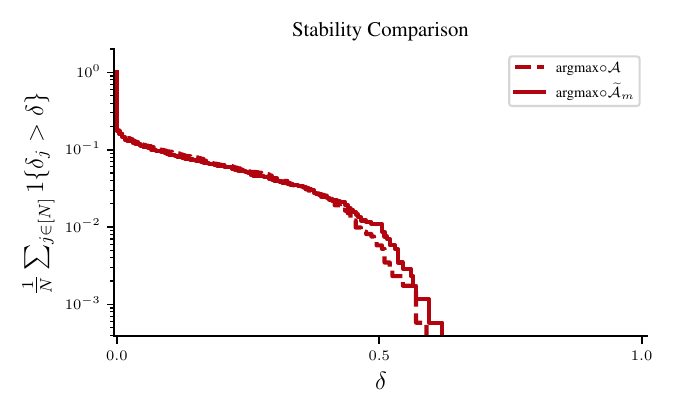}
        \caption{LDA, $k=1000$}
        \label{fig:k-grid-lda-1000}
    \end{subfigure}
    \hfill
    \begin{subfigure}{0.46\textwidth}
        \centering
        \includegraphics[width=\textwidth]{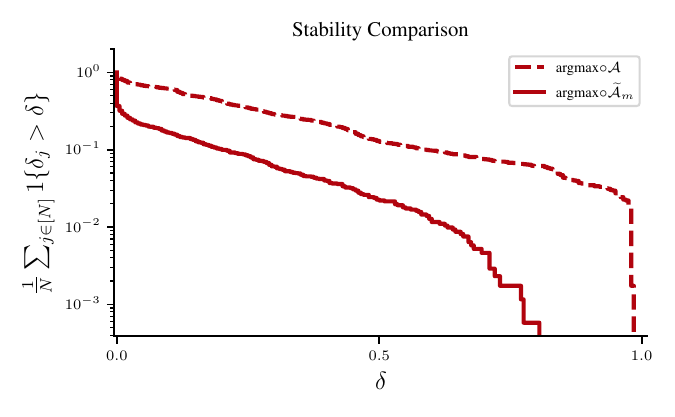}
        \caption{Decision Tree, $k=1000$}
        \label{fig:k-grid-dt-1000}
    \end{subfigure}

    \caption{Instability plots for jackknife-stable classifiers under leave-$k$-out perturbations with $B=300$ at increasing $k$. The curves show the proportion of samples in the test set with instability greater than $\delta$ for unbagged (dashed) and subbagged (solid) algorithms. This illustrates the different behaviors as the perturbation level grows, where the effect of subbagging can become negligible or flip to become beneficial.} 
    \label{fig:leave-k-out-grid}
\end{figure}

\begin{figure}[H]
    \centering
    \begin{subfigure}{0.45\textwidth}
        \centering
        \includegraphics[width=\textwidth]{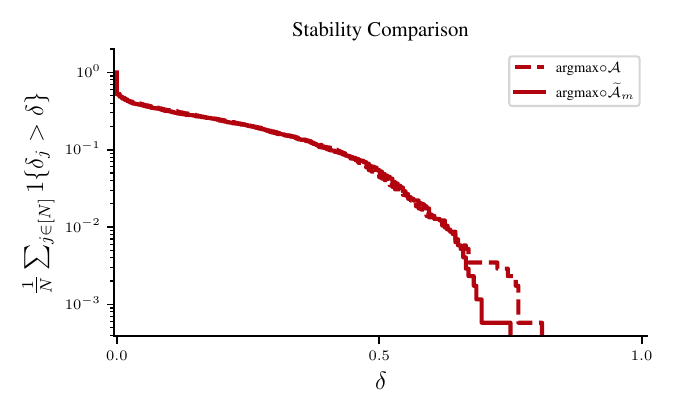}
        \caption{Naive Bayes}
        \label{fig:app-bs-left}
    \end{subfigure}
    \hfill
    \begin{subfigure}{0.45\textwidth}
        \centering
        \includegraphics[width=\textwidth]{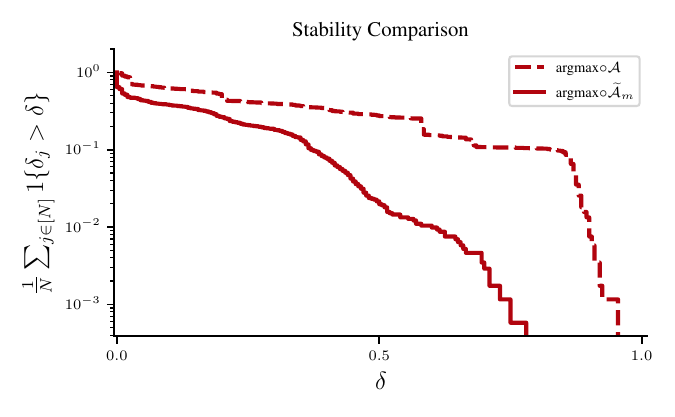}
        \caption{Regularized Decision Tree}
        \label{fig:app-bs-right}
    \end{subfigure}
    \caption{Instability plots for jackknife-stable classifiers with bootstrap perturbations and $B=300$. The curves show the proportion of samples in the test set with instability greater than $\delta$ for unbagged (dashed) and subbagged (solid) algorithms.}
    \label{fig:app-bs-instability}
\end{figure}

\begin{figure}[H]
    \centering
    \begin{subfigure}{0.45\textwidth}
        \centering
        \includegraphics[width=\textwidth]{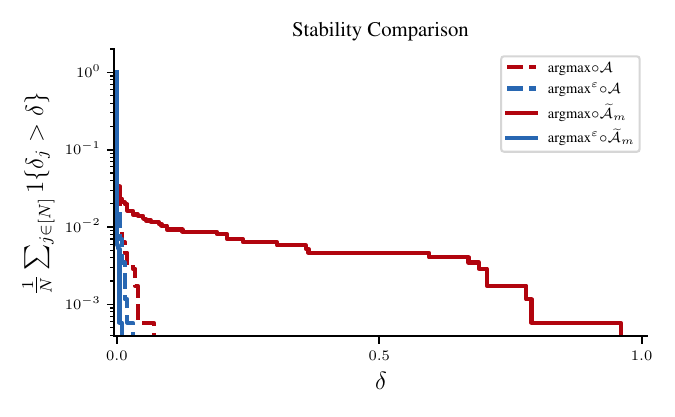}
        \caption{Leave-one-out perturbation}
        \label{fig:ins-plot-perturbed}
    \end{subfigure}
    \hfill
    \begin{subfigure}{0.45\textwidth}
        \centering
        \includegraphics[width=\textwidth]{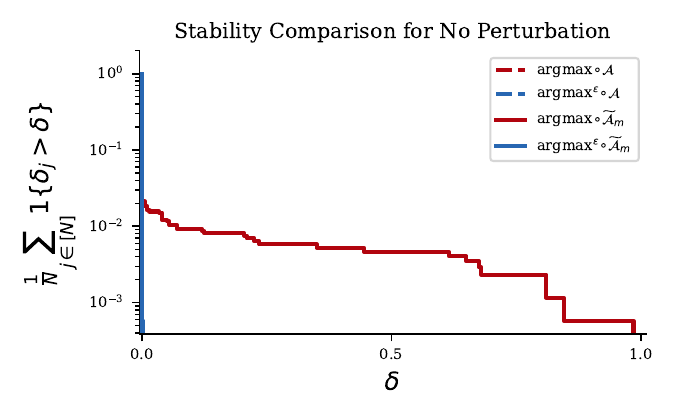}
        \caption{No perturbation}
        \label{fig:ins-plot-unperturbed}
    \end{subfigure}
    \caption{Instability plots with and without perturbations for the naive Bayes classifier and $B=2500$. The curves show the proportion of samples in the test set with instability greater than $\delta$. The increase in instability with subbagging is present even without any perturbation.}
    \label{fig:instability-plots-no-perturb}
\end{figure}

\subsubsection{Rashomon}
\label{app:rashomon-figures}

\begin{figure}[H]
    \centering
    \begin{subfigure}{0.45\textwidth}
        \centering
        \includegraphics[width=\textwidth]{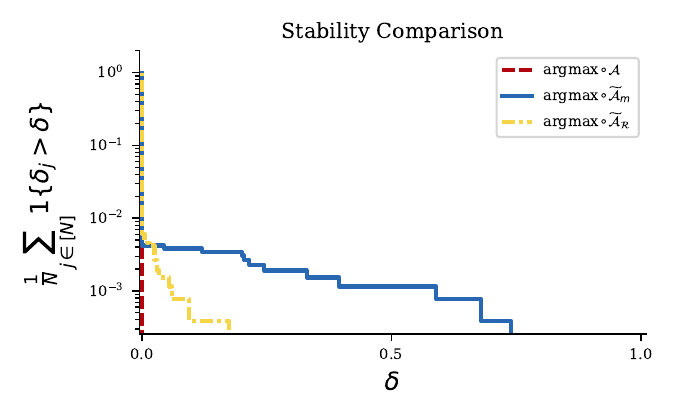}
        \caption{HELOC Dataset, $n$ = 7844}
        \label{fig:ins-rashomon-heloc-loo}
    \end{subfigure}
    \hfill
    \begin{subfigure}{0.45\textwidth}
        \centering
        \includegraphics[width=\textwidth]{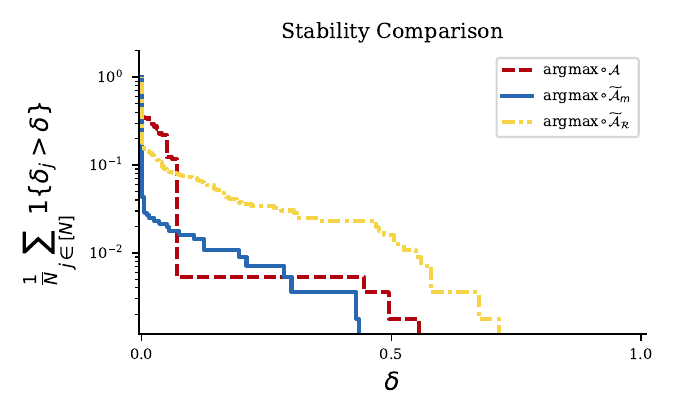}
        \caption{Synthetic Dataset, $n$ = 204}
        \label{fig:ins-rashomon-synth-loo}
    \end{subfigure}
    \caption{Instability plots from leave-one-out perturbations with the Rashomon ensembles. The curves show the proportion of samples in the test set with instability greater than $\delta$.}
    \label{fig:rashomon-plots-loo}
\end{figure}

\begin{figure}[H]
    \centering
    \begin{subfigure}{0.48\textwidth}
        \centering
        \includegraphics[width=\textwidth]{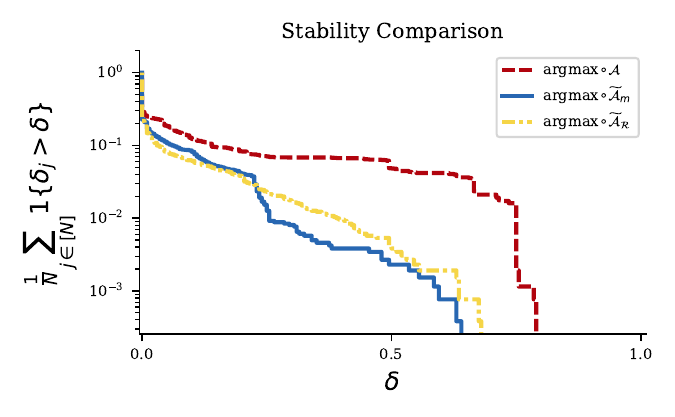}
        \caption{HELOC Dataset, $n$ = 7844}
        \label{fig:ins-rashomon-1}
    \end{subfigure}
    \hfill
    \begin{subfigure}{0.48\textwidth}
        \centering
        \includegraphics[width=\textwidth]{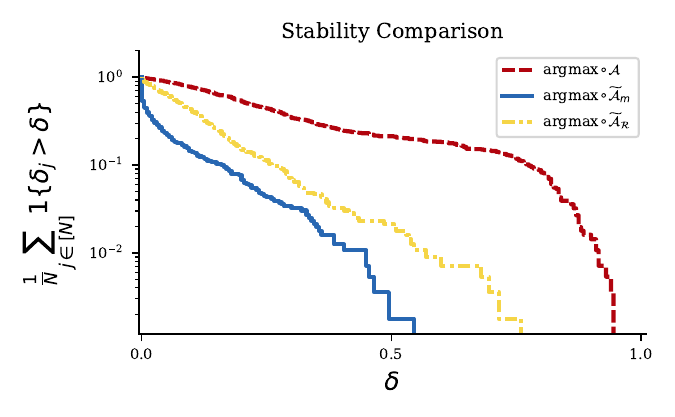}
        \caption{Synthetic Dataset, $n$ = 204}
        \label{fig:ins-rashomon-2}
    \end{subfigure}
    \caption{Instability plots from bootstrap perturbations with the Rashomon ensembles. The curves show the proportion of samples in the test set with instability greater than $\delta$.}
    \label{fig:rashomon-plots-bs}
\end{figure}

\begin{figure}[H]
    \centering
    \begin{subfigure}{0.45\textwidth}
        \centering
        \includegraphics[width=\textwidth]{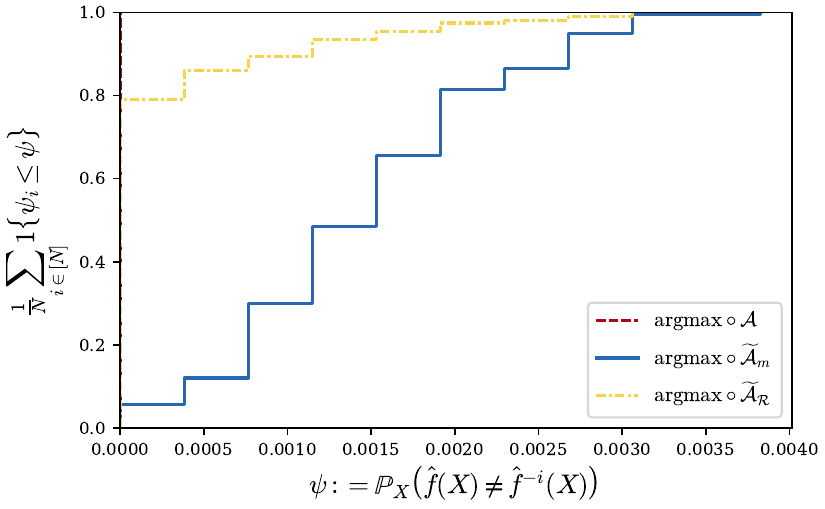}
        \caption{HELOC Dataset}
        \label{fig:infl-rashomon-1}
    \end{subfigure}
    \hfill
    \begin{subfigure}{0.45\textwidth}
        \centering
        \includegraphics[width=\textwidth]{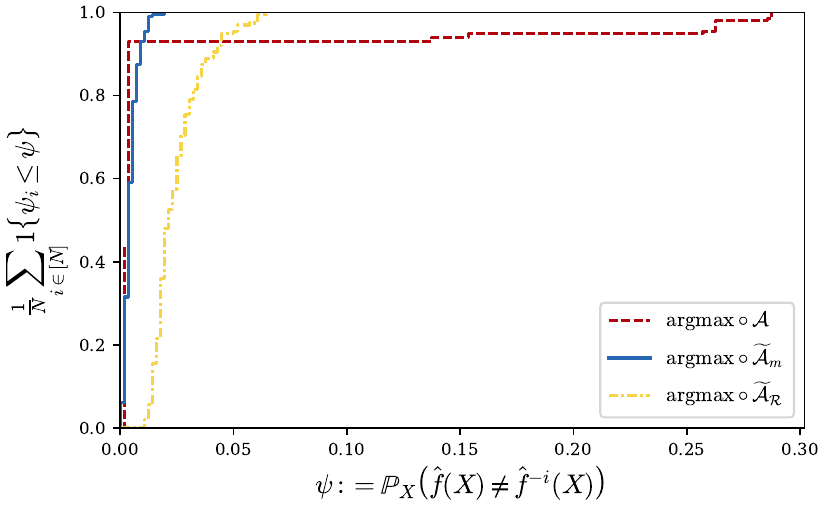}
        \caption{Synthetic Dataset}
        \label{fig:infl-rashomon-2}
    \end{subfigure}
    \caption{Empirical cumulative distribution function of influence as defined by $\mathbb{P}_X(\hat{f}(X) \neq \hat{f}^{\setminus i}(X))$ for unbagged (red dashed), subbagged (blue solid), and Rashomon ensemble (yellow solid) algorithms for leave-one-out perturbations. The Rashomon ensemble generates an equalization effect similar to that of subbagging. However, this does appear to break down for the HELOC example that is fully jackknife stable, though it may be that this jackknife definition of influence is simply a poor proxy in this scenario.}
    \label{fig:rashomon-influence-plots}
\end{figure}

\begin{figure}[H]
    \centering
    \begin{subfigure}{0.45\textwidth}
        \centering
        \includegraphics[width=\textwidth]{Figures/instability_plot_heloc_loo.pdf}
        \caption{Leave-one-out perturbation}
        \label{fig:ins-rashomon-yes-perturb}
    \end{subfigure}
    \hfill
    \begin{subfigure}{0.45\textwidth}
        \centering
        \includegraphics[width=\textwidth]{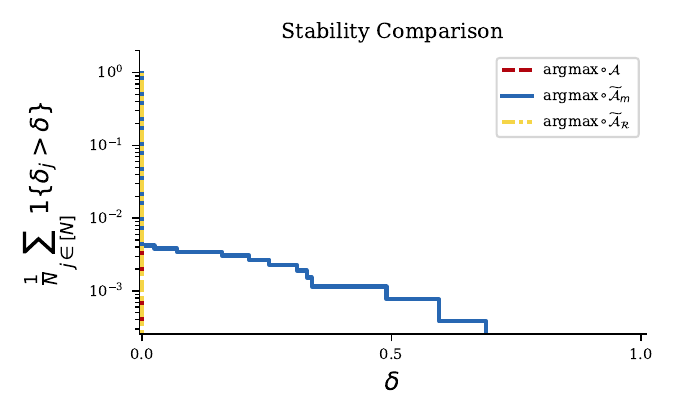}
        \caption{No perturbation}
        \label{fig:ins-rashomon-no-perturb}
    \end{subfigure}
    \caption{Instability plots with and without perturbations with the Rashomon ensembles on the HELOC dataset. The curves show the proportion of samples in the test set with instability greater than $\delta$. The Rashomon ensemble is consistent without perturbation.}
    \label{fig:rashomon-plots-no-perturb}
\end{figure}

\subsubsection{Sensitivity Analysis}

\begin{figure}[H]
    \centering
    \begin{subfigure}{0.48\textwidth}
        \centering
        \includegraphics[width=\textwidth]{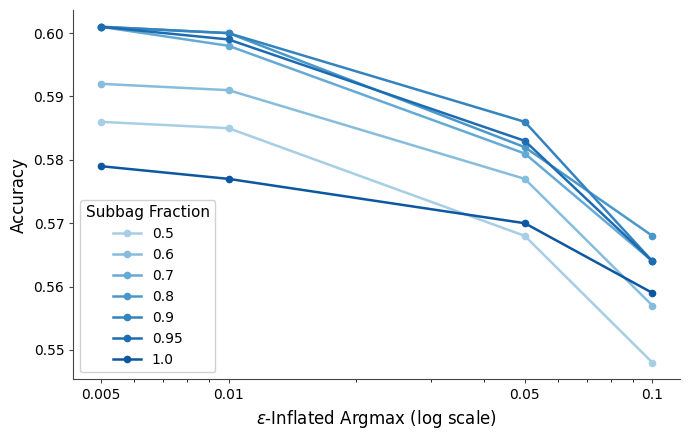}
        \caption{}
        \label{fig:ins-tl}
    \end{subfigure}
    \hfill
    \begin{subfigure}{0.48\textwidth}
        \centering
        \includegraphics[width=\textwidth]{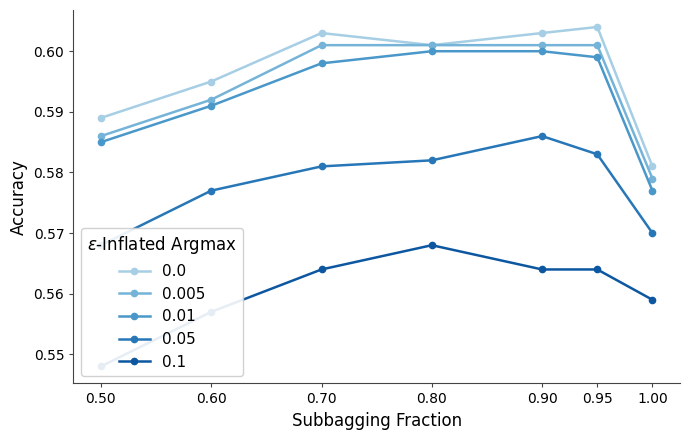}
        \caption{}
        \label{fig:ins-tr}
    \end{subfigure}

    \vspace{1em}

    \begin{subfigure}{0.48\textwidth}
        \centering
        \includegraphics[width=\textwidth]{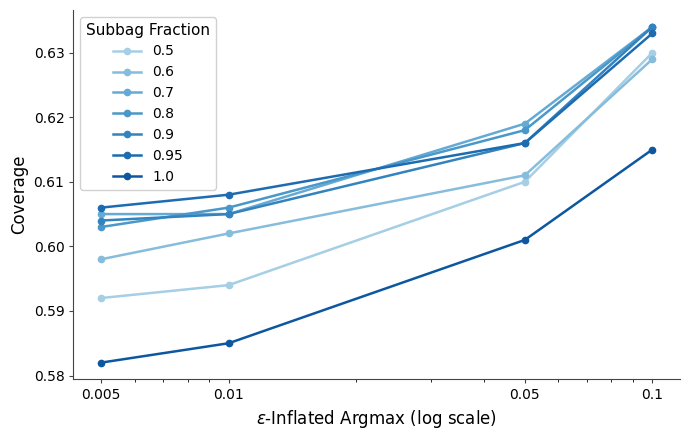}
        \caption{}
        \label{fig:ins-bl}
    \end{subfigure}
    \hfill
    \begin{subfigure}{0.48\textwidth}
        \centering
        \includegraphics[width=\textwidth]{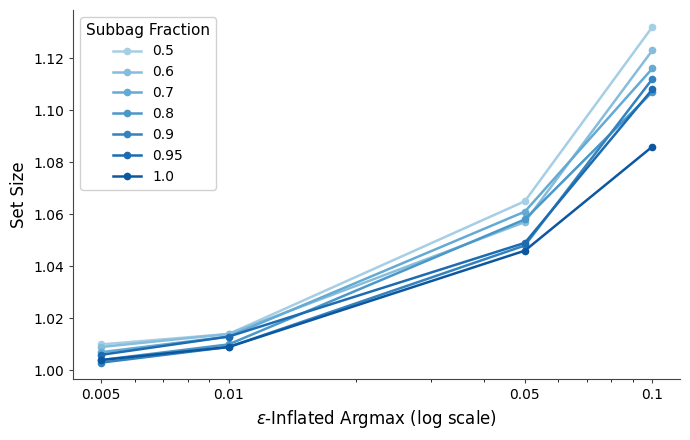}
        \caption{}
        \label{fig:ins-br}
    \end{subfigure}

    \caption{Subbagging fraction ($\frac{m}{n}$) and $\varepsilon$-inflated argmax sensitivity analysis for additional metrics. The results illustrate how the subbagging fraction and the $\varepsilon$-inflated argmax can affect accuracy, coverage, and set size. As we increase the $\varepsilon$-inflated argmax, we always trade off accuracy for coverage and increase set size. Subbagging can moderate these effects and is known to have an effect on accuracy, either positive or negative depending on the classifier.}
    \label{fig:sensitivity-plots-extra}
\end{figure}

\subsection{Tables}
\label{app:tables}

We can also measure the output stability more directly by considering the following.

\begin{equation}
\beta_{\text{eucl-instability}} := \frac{1}{N} \sum_{j=1}^{N} \frac{1}{T} \sum_{t=1}^{T} \left\| \hat{p}(\tilde{X}_j) - \hat{p}^{\prime}_t(\tilde{X}_j) \right\| \, ,
\end{equation}

which is the average Euclidean distance between the prediction probability vector from the model trained on the full dataset and that of the model trained on each perturbed dataset. 

In our experiments, this output instability roughly mirrored the selection stability, although this will not always be the case. It may be that an ensemble technique has little effect on the Euclidean instability but a noticeable effect on the selection instability. In these cases, since the technique acts before the selection rule, any improvement in selection instability must be viewed somewhat skeptically because it is unlikely to be a reliable or consistent effect.

Standard errors are computed across the $N$ test-set points. For each summary metric $\beta = \frac{1}{N}\sum_{j=1}^N g_j$, where $g_j$ is the metric for a single sample in the test set, we report $\mathrm{SE}(\beta) = \mathrm{sd}(\{g_j\})/\sqrt{N}$ as the standard error of the test-set mean. This does not account for Monte Carlo variability from the finite perturbations, so it should be viewed as a lower bound on the total uncertainty.

\subsubsection{Unstable Classifiers with Leave-One-Out Perturbations}

\rowcolors{2}{gray!25}{white}
\begin{table}[H]
\centering
\caption{Summary results from the verbal autopsy dataset and Multi-Layer Perceptron (MLP) classifier validated on Dar, Tanzania holdout data with inflated argmax $\epsilon = 0.05$. Standard errors are shown in parentheses and arrow direction denotes higher/lower values are desired.}
\resizebox{\textwidth}{!}{%
    {\renewcommand{\arraystretch}{1.5} 
    \begin{tabular}{c c c c c c c c}
    \rowcolor{gray!60}
    Sel. rule & Algo & $\beta_{\text{accuracy}}\nearrow$ & $\beta_{\text{coverage}}\nearrow$ & $\beta_{\text{eucl-instability}}\searrow$ & $\beta_{\text{mean-instability}}\searrow$ & $\beta_{\text{max-instability}}\searrow$ & $\beta_{\text{size}}\searrow$ \\\hline
	 argmax & $\mathcal{A}$ & 0.581 (0.012) & 0.581 (0.012) & 0.156 (0.003) & 0.155 (0.006) & 0.986 & 1.000 (0.000) \\
	 argmax & $\widetilde{\mathcal{A}}_m$ & 0.590 (0.012) & 0.590 (0.012) & 0.010 (0.000) & 0.011 (0.002) & 0.908 & 1.000 (0.000) \\
	 argmax$^\varepsilon$ & $\mathcal{A}$ & 0.570 (0.012) & 0.601 (0.012) & 0.156 (0.003) & 0.122 (0.005) & 0.952 & 1.046 (0.005) \\
	 argmax$^\varepsilon$ & $\widetilde{\mathcal{A}}_m$ & 0.570 (0.012) & 0.610 (0.012) & 0.010 (0.000) & 0.000 (0.000) & 0.022 & 1.064 (0.006) \\

\end{tabular}}
    }
    \medskip
    \label{tab:mlp}
\end{table}

\rowcolors{2}{gray!25}{white}
\begin{table}[H]
\centering
\caption{Summary results from the verbal autopsy dataset and gradient boosting classifier validated on Dar, Tanzania holdout data with inflated argmax $\epsilon = 0.05$. Standard errors are shown in parentheses and arrow direction denotes higher/lower values are desired.}
\resizebox{\textwidth}{!}{%
    {\renewcommand{\arraystretch}{1.5} 
    \begin{tabular}{c c c c c c c c}
    \rowcolor{gray!60}
    Sel. rule & Algo & $\beta_{\text{accuracy}}\nearrow$ & $\beta_{\text{coverage}}\nearrow$ & $\beta_{\text{eucl-instability}}\searrow$ & $\beta_{\text{mean-instability}}\searrow$ & $\beta_{\text{max-instability}}\searrow$ & $\beta_{\text{size}}\searrow$ \\\hline
	 argmax & $\mathcal{A}$ & 0.542 (0.012) & 0.542 (0.012) & 0.080 (0.001) & 0.159 (0.008) & 1.000 & 1.000 (0.000) \\
	 argmax & $\widetilde{\mathcal{A}}_m$ & 0.568 (0.012) & 0.568 (0.012) & 0.002 (0.000) & 0.003 (0.001) & 0.826 & 1.000 (0.000) \\
	 argmax$^\varepsilon$ & $\mathcal{A}$ & 0.483 (0.012) & 0.594 (0.012) & 0.080 (0.001) & 0.068 (0.005) & 1.000 & 1.203 (0.011) \\
	 argmax$^\varepsilon$ & $\widetilde{\mathcal{A}}_m$ & 0.505 (0.012) & 0.617 (0.012) & 0.002 (0.000) & 0.000 (0.000) & 0.000 & 1.155 (0.009) \\

\end{tabular}}
    }
    \medskip
    \label{tab:gb}
\end{table}

\subsubsection{Stable Classifiers with Leave-One-Out Perturbations}

\rowcolors{2}{gray!25}{white}
\begin{table}[H]
\centering
\caption{Summary results from the verbal autopsy dataset and InSilicoVA \citep{mccormick2016insilicova} classifier validated on Dar, Tanzania holdout data with inflated argmax $\epsilon = 0.1$ --- a lower $\epsilon$ produced negligible inflation, because the model makes highly confident predictions, so we must increase $\epsilon$ to achieve a comparable effect. Standard errors are shown in parentheses and arrow direction denotes higher/lower values are desired.}
\resizebox{\textwidth}{!}{%
    {\renewcommand{\arraystretch}{1.5} 
    \begin{tabular}{c c c c c c c c}
    \rowcolor{gray!60}
    Sel. rule & Algo & $\beta_{\text{accuracy}}\nearrow$ & $\beta_{\text{coverage}}\nearrow$ & $\beta_{\text{eucl-instability}}\searrow$ & $\beta_{\text{mean-instability}}\searrow$ & $\beta_{\text{max-instability}}\searrow$ & $\beta_{\text{size}}\searrow$ \\\hline
	 argmax & $\mathcal{A}$ & 0.541 (0.012) & 0.541 (0.012) & 0.003 (0.000) & 0.002 (0.000) & 0.150 & 1.000 (0.000) \\
	 argmax & $\widetilde{\mathcal{A}}_m$ & 0.548 (0.012) & 0.548 (0.012) & 0.018 (0.000) & 0.013 (0.002) & 1.000 & 1.000 (0.000) \\
	 argmax$^\varepsilon$ & $\mathcal{A}$ & 0.525 (0.012) & 0.552 (0.012) & 0.003 (0.000) & 0.001 (0.000) & 0.150 & 1.040 (0.005) \\
	 argmax$^\varepsilon$ & $\widetilde{\mathcal{A}}_m$ & 0.536 (0.012) & 0.568 (0.012) & 0.018 (0.000) & 0.000 (0.000) & 0.045 & 1.045 (0.005) \\

\end{tabular}}
    }
    \medskip
    \label{tab:openva-res}
\end{table}

\rowcolors{2}{gray!25}{white}
\begin{table}[H]
\centering
\caption{Summary results from the verbal autopsy dataset and decision tree classifier validated on Dar, Tanzania holdout data with inflated argmax $\epsilon = 0.05$. Standard errors are shown in parentheses and arrow direction denotes higher/lower values are desired.}
\resizebox{\textwidth}{!}{%
    {\renewcommand{\arraystretch}{1.5} 
    \begin{tabular}{c c c c c c c c}
    \rowcolor{gray!60}
    Sel. rule & Algo & $\beta_{\text{accuracy}}\nearrow$ & $\beta_{\text{coverage}}\nearrow$ & $\beta_{\text{eucl-instability}}\searrow$ & $\beta_{\text{mean-instability}}\searrow$ & $\beta_{\text{max-instability}}\searrow$ & $\beta_{\text{size}}\searrow$ \\\hline
	 argmax & $\mathcal{A}$ & 0.568 (0.012) & 0.568 (0.012) & 0.004 (0.000) & 0.002 (0.000) & 0.158 & 1.000 (0.000) \\
	 argmax & $\widetilde{\mathcal{A}}_m$ & 0.547 (0.012) & 0.547 (0.012) & 0.018 (0.000) & 0.013 (0.002) & 0.914 & 1.000 (0.000) \\
	 argmax$^\varepsilon$ & $\mathcal{A}$ & 0.564 (0.012) & 0.570 (0.012) & 0.004 (0.000) & 0.001 (0.000) & 0.158 & 1.010 (0.003) \\
	 argmax$^\varepsilon$ & $\widetilde{\mathcal{A}}_m$ & 0.535 (0.012) & 0.565 (0.012) & 0.018 (0.000) & 0.000 (0.000) & 0.190 & 1.035 (0.005) \\

\end{tabular}}
    }
    \medskip
    \label{tab:tree}
\end{table}

\rowcolors{2}{gray!25}{white}
\begin{table}[H]
\centering
\caption{Summary results from the verbal autopsy dataset and naive Bayes classifier validated on Dar, Tanzania holdout data with inflated argmax $\epsilon = 0.05$. Standard errors are shown in parentheses and arrow direction denotes higher/lower values are desired.}
\resizebox{\textwidth}{!}{%
    {\renewcommand{\arraystretch}{1.5} 
    \begin{tabular}{c c c c c c c c}
    \rowcolor{gray!60}
    Sel. rule & Algo & $\beta_{\text{accuracy}}\nearrow$ & $\beta_{\text{coverage}}\nearrow$ & $\beta_{\text{eucl-instability}}\searrow$ & $\beta_{\text{mean-instability}}\searrow$ & $\beta_{\text{max-instability}}\searrow$ & $\beta_{\text{size}}\searrow$ \\\hline
	 argmax & $\mathcal{A}$ & 0.517 (0.012) & 0.517 (0.012) & 0.001 (0.000) & 0.000 (0.000) & 0.082 & 1.000 (0.000) \\
	 argmax & $\widetilde{\mathcal{A}}_m$ & 0.516 (0.012) & 0.516 (0.012) & 0.012 (0.000) & 0.012 (0.002) & 0.980 & 1.000 (0.000) \\
	 argmax$^\varepsilon$ & $\mathcal{A}$ & 0.517 (0.012) & 0.517 (0.012) & 0.001 (0.000) & 0.000 (0.000) & 0.024 & 1.001 (0.001) \\
	 argmax$^\varepsilon$ & $\widetilde{\mathcal{A}}_m$ & 0.509 (0.012) & 0.523 (0.012) & 0.012 (0.000) & 0.000 (0.000) & 0.182 & 1.027 (0.004) \\

\end{tabular}}
    }
    \medskip
    \label{tab:nb}
\end{table}

\rowcolors{2}{gray!25}{white}
\begin{table}[H]
\centering
\caption{Summary results from the verbal autopsy dataset and K-Nearest Neighbors (KNN) classifier validated on Dar, Tanzania holdout data with inflated argmax $\epsilon = 0.05$. Standard errors are shown in parentheses and arrow direction denotes higher/lower values are desired.}
\resizebox{\textwidth}{!}{%
    {\renewcommand{\arraystretch}{1.5} 
    \begin{tabular}{c c c c c c c c}
    \rowcolor{gray!60}
    Sel. rule & Algo & $\beta_{\text{accuracy}}\nearrow$ & $\beta_{\text{coverage}}\nearrow$ & $\beta_{\text{eucl-instability}}\searrow$ & $\beta_{\text{mean-instability}}\searrow$ & $\beta_{\text{max-instability}}\searrow$ & $\beta_{\text{size}}\searrow$ \\\hline
	 argmax & $\mathcal{A}$ & 0.385 (0.012) & 0.385 (0.012) & 0.000 (0.000) & 0.000 (0.000) & 0.284 & 1.000 (0.000) \\
	 argmax & $\widetilde{\mathcal{A}}_m$ & 0.395 (0.012) & 0.395 (0.012) & 0.011 (0.000) & 0.019 (0.002) & 0.976 & 1.000 (0.000) \\
	 argmax$^\varepsilon$ & $\mathcal{A}$ & 0.336 (0.011) & 0.432 (0.012) & 0.000 (0.000) & 0.000 (0.000) & 0.223 & 1.215 (0.011) \\
	 argmax$^\varepsilon$ & $\widetilde{\mathcal{A}}_m$ & 0.368 (0.012) & 0.421 (0.012) & 0.011 (0.000) & 0.000 (0.000) & 0.028 & 1.114 (0.008) \\

\end{tabular}}
    }
    \medskip
    \label{tab:knn}
\end{table}

\rowcolors{2}{gray!25}{white}
\begin{table}[H]
\centering
\caption{Summary results from the verbal autopsy dataset and Linear Discriminant Analysis (LDA) classifier validated on Dar, Tanzania holdout data with inflated argmax $\epsilon = 0.05$. Standard errors are shown in parentheses and arrow direction denotes higher/lower values are desired.}
\resizebox{\textwidth}{!}{%
    {\renewcommand{\arraystretch}{1.5} 
    \begin{tabular}{c c c c c c c c}
    \rowcolor{gray!60}
    Sel. rule & Algo & $\beta_{\text{accuracy}}\nearrow$ & $\beta_{\text{coverage}}\nearrow$ & $\beta_{\text{eucl-instability}}\searrow$ & $\beta_{\text{mean-instability}}\searrow$ & $\beta_{\text{max-instability}}\searrow$ & $\beta_{\text{size}}\searrow$ \\\hline
	 argmax & $\mathcal{A}$ & 0.645 (0.012) & 0.645 (0.012) & 0.001 (0.000) & 0.001 (0.000) & 0.280 & 1.000 (0.000) \\
	 argmax & $\widetilde{\mathcal{A}}_m$ & 0.646 (0.012) & 0.646 (0.012) & 0.005 (0.000) & 0.005 (0.001) & 0.958 & 1.000 (0.000) \\
	 argmax$^\varepsilon$ & $\mathcal{A}$ & 0.635 (0.012) & 0.655 (0.011) & 0.001 (0.000) & 0.000 (0.000) & 0.004 & 1.028 (0.004) \\
	 argmax$^\varepsilon$ & $\widetilde{\mathcal{A}}_m$ & 0.638 (0.012) & 0.657 (0.011) & 0.005 (0.000) & 0.000 (0.000) & 0.005 & 1.030 (0.004) \\

\end{tabular}}
    }
    \medskip
    \label{tab:lda}
\end{table}

\rowcolors{2}{gray!25}{white}
\begin{table}[H]
\centering
\caption{Summary results from the verbal autopsy dataset and logistic regression classifier validated on Dar, Tanzania holdout data with inflated argmax $\epsilon = 0.05$. Standard errors are shown in parentheses and arrow direction denotes higher/lower values are desired.}
\resizebox{\textwidth}{!}{%
    {\renewcommand{\arraystretch}{1.5} 
    \begin{tabular}{c c c c c c c c}
    \rowcolor{gray!60}
    Sel. rule & Algo & $\beta_{\text{accuracy}}\nearrow$ & $\beta_{\text{coverage}}\nearrow$ & $\beta_{\text{eucl-instability}}\searrow$ & $\beta_{\text{mean-instability}}\searrow$ & $\beta_{\text{max-instability}}\searrow$ & $\beta_{\text{size}}\searrow$ \\\hline
	 argmax & $\mathcal{A}$ & 0.593 (0.012) & 0.593 (0.012) & 0.001 (0.000) & 0.001 (0.000) & 0.210 & 1.000 (0.000) \\
	 argmax & $\widetilde{\mathcal{A}}_m$ & 0.604 (0.012) & 0.604 (0.012) & 0.006 (0.000) & 0.005 (0.001) & 0.720 & 1.000 (0.000) \\
	 argmax$^\varepsilon$ & $\mathcal{A}$ & 0.584 (0.012) & 0.609 (0.012) & 0.001 (0.000) & 0.000 (0.000) & 0.004 & 1.044 (0.005) \\
	 argmax$^\varepsilon$ & $\widetilde{\mathcal{A}}_m$ & 0.593 (0.012) & 0.625 (0.012) & 0.006 (0.000) & 0.000 (0.000) & 0.285 & 1.053 (0.006) \\

\end{tabular}}
    }
    \medskip
    \label{tab:lr}
\end{table}

\rowcolors{2}{gray!25}{white}
\begin{table}[!htbp]
\centering
\caption{Summary results from the verbal autopsy dataset and a heavily regularized decision tree classifier validated on Dar, Tanzania holdout data with inflated argmax $\epsilon = 0.05$. Standard errors are shown in parentheses and arrow direction denotes higher/lower values are desired.}
\resizebox{\textwidth}{!}{%
    {\renewcommand{\arraystretch}{1.5} 
    \begin{tabular}{c c c c c c c c}
    \rowcolor{gray!60}
    Sel. rule & Algo & $\beta_{\text{accuracy}}\nearrow$ & $\beta_{\text{coverage}}\nearrow$ & $\beta_{\text{eucl-instability}}\searrow$ & $\beta_{\text{mean-instability}}\searrow$ & $\beta_{\text{max-instability}}\searrow$ & $\beta_{\text{size}}\searrow$ \\\hline
	 argmax & $\mathcal{A}$ & 0.414 (0.012) & 0.414 (0.012) & 0.000 (0.000) & 0.000 (0.000) & 0.004 & 1.000 (0.000) \\
	 argmax & $\widetilde{\mathcal{A}}_m$ & 0.499 (0.012) & 0.499 (0.012) & 0.010 (0.000) & 0.003 (0.001) & 0.794 & 1.000 (0.000) \\
	 argmax$^\varepsilon$ & $\mathcal{A}$ & 0.409 (0.012) & 0.420 (0.012) & 0.000 (0.000) & 0.000 (0.000) & 0.004 & 1.011 (0.003) \\
	 argmax$^\varepsilon$ & $\widetilde{\mathcal{A}}_m$ & 0.496 (0.012) & 0.509 (0.012) & 0.010 (0.000) & 0.000 (0.000) & 0.000 & 1.016 (0.003) \\

\end{tabular}}
    }
    \medskip
    \label{tab:tree-lc}
\end{table}

\subsubsection{Rashomon}

\rowcolors{2}{gray!25}{white}
\begin{table}[H]
\centering
\caption{Summary results from the synthetic dataset and SORTeD decision tree classifier with inflated argmax $\epsilon = 0.1$ and leave-25-out perturbations. On the full training dataset the Rashomon set had 6,573 trees, of which 1,373 produced unique prediction vectors, and those were clustered down to 10. Standard errors are shown in parentheses and arrow direction denotes higher/lower values are desired.}
\resizebox{\textwidth}{!}{%
    {\renewcommand{\arraystretch}{1.5} 
    \begin{tabular}{c c c c c c c c}
    \rowcolor{gray!60}
    Sel. rule & Algo & $\beta_{\text{accuracy}}\nearrow$ & $\beta_{\text{coverage}}\nearrow$ & $\beta_{\text{eucl-instability}}\searrow$ & $\beta_{\text{mean-instability}}\searrow$ & $\beta_{\text{max-instability}}\searrow$ & $\beta_{\text{size}}\searrow$ \\\hline
	 argmax & $\mathcal{A}$ & 0.779 (0.018) & 0.779 (0.018) & 0.384 (0.013) & 0.275 (0.011) & 0.960 & 1.000 (0.000) \\
	 argmax & $\widetilde{\mathcal{A}}_m$ & 0.975 (0.007) & 0.975 (0.007) & 0.064 (0.001) & 0.033 (0.005) & 0.860 & 1.000 (0.000) \\
	 argmax & $\widetilde{\mathcal{A}}_{\mathcal{R}}$ & 0.968 (0.007) & 0.968 (0.007) & 0.181 (0.004) & 0.080 (0.005) & 0.690 & 1.000 (0.000) \\
	 argmax$^\varepsilon$ & $\mathcal{A}$ & 0.779 (0.018) & 0.779 (0.018) & 0.384 (0.013) & 0.275 (0.011) & 0.960 & 1.000 (0.000) \\
	 argmax$^\varepsilon$ & $\widetilde{\mathcal{A}}_m$ & 0.966 (0.008) & 0.979 (0.006) & 0.064 (0.001) & 0.008 (0.002) & 0.500 & 1.012 (0.005) \\
	 argmax$^\varepsilon$ & $\widetilde{\mathcal{A}}_{\mathcal{R}}$ & 0.930 (0.011) & 0.968 (0.007) & 0.181 (0.004) & 0.066 (0.004) & 0.570 & 1.038 (0.008) \\

\end{tabular}}
    }
    \medskip
    \label{tab:rashomon-1}
\end{table}

\rowcolors{2}{gray!25}{white}
\begin{table}[!htbp]
\centering
\caption{Summary results from the FICO HELOC dataset and SORTeD decision tree classifier with inflated argmax $\epsilon = 0.1$ and leave-1000-out perturbations. On the full training set, the Rashomon set contained 5,497 trees of which 347 produced unique prediction vectors, and those were clustered down to 50. Standard errors are shown in parentheses and arrow direction denotes higher/lower values are desired.}
\resizebox{\textwidth}{!}{%
    {\renewcommand{\arraystretch}{1.5} 
    \begin{tabular}{c c c c c c c c}
    \rowcolor{gray!60}
    Sel. rule & Algo & $\beta_{\text{accuracy}}\nearrow$ & $\beta_{\text{coverage}}\nearrow$ & $\beta_{\text{eucl-instability}}\searrow$ & $\beta_{\text{mean-instability}}\searrow$ & $\beta_{\text{max-instability}}\searrow$ & $\beta_{\text{size}}\searrow$ \\\hline
	 argmax & $\mathcal{A}$ & 0.858 (0.007) & 0.858 (0.007) & 0.063 (0.002) & 0.041 (0.003) & 0.590 & 1.000 (0.000) \\
	 argmax & $\widetilde{\mathcal{A}}_m$ & 0.857 (0.007) & 0.857 (0.007) & 0.016 (0.000) & 0.005 (0.001) & 0.555 & 1.000 (0.000) \\
	 argmax & $\widetilde{\mathcal{A}}_{\mathcal{R}}$ & 0.858 (0.007) & 0.858 (0.007) & 0.020 (0.001) & 0.005 (0.001) & 0.615 & 1.000 (0.000) \\
	 argmax$^\varepsilon$ & $\mathcal{A}$ & 0.858 (0.007) & 0.858 (0.007) & 0.063 (0.002) & 0.041 (0.003) & 0.590 & 1.000 (0.000) \\
	 argmax$^\varepsilon$ & $\widetilde{\mathcal{A}}_m$ & 0.854 (0.007) & 0.859 (0.007) & 0.016 (0.000) & 0.001 (0.000) & 0.105 & 1.004 (0.001) \\
	 argmax$^\varepsilon$ & $\widetilde{\mathcal{A}}_{\mathcal{R}}$ & 0.857 (0.007) & 0.860 (0.007) & 0.020 (0.001) & 0.002 (0.000) & 0.400 & 1.003 (0.001) \\

\end{tabular}}
    }
    \medskip
    \label{tab:results-extended}
\end{table}

\rowcolors{2}{gray!25}{white}
\begin{table}[H]
\centering
\caption{Summary results from the COMPAS Recidivism dataset \citep{compas2016} and SORTeD decision tree classifier with inflated argmax $\epsilon = 0.1$ and bootstrap perturbations. On the full training dataset the Rashomon set had 3,660 trees, of which 340 produced unique prediction vectors, and those were clustered down to 50. Standard errors are shown in parentheses and arrow direction denotes higher/lower values are desired.}
\resizebox{\textwidth}{!}{%
    {\renewcommand{\arraystretch}{1.5} 
    \begin{tabular}{c c c c c c c c}
    \rowcolor{gray!60}
    Sel. rule & Algo & $\beta_{\text{accuracy}}\nearrow$ & $\beta_{\text{coverage}}\nearrow$ & $\beta_{\text{eucl-instability}}\searrow$ & $\beta_{\text{mean-instability}}\searrow$ & $\beta_{\text{max-instability}}\searrow$ & $\beta_{\text{size}}\searrow$ \\\hline
	 argmax & $\mathcal{A}$ & 0.664 (0.013) & 0.664 (0.013) & 0.070 (0.002) & 0.091 (0.004) & 0.620 & 1.000 (0.000) \\
	 argmax & $\widetilde{\mathcal{A}}_m$ & 0.669 (0.013) & 0.669 (0.013) & 0.042 (0.001) & 0.067 (0.003) & 0.570 & 1.000 (0.000) \\
	 argmax & $\widetilde{\mathcal{A}}_{\mathcal{R}}$ & 0.662 (0.013) & 0.662 (0.013) & 0.075 (0.002) & 0.052 (0.004) & 0.690 & 1.000 (0.000) \\
	 argmax$^\varepsilon$ & $\mathcal{A}$ & 0.664 (0.013) & 0.664 (0.013) & 0.070 (0.002) & 0.091 (0.004) & 0.620 & 1.000 (0.000) \\
	 argmax$^\varepsilon$ & $\widetilde{\mathcal{A}}_m$ & 0.660 (0.013) & 0.676 (0.013) & 0.042 (0.001) & 0.035 (0.002) & 0.270 & 1.016 (0.004) \\
	 argmax$^\varepsilon$ & $\widetilde{\mathcal{A}}_{\mathcal{R}}$ & 0.657 (0.014) & 0.666 (0.013) & 0.075 (0.002) & 0.040 (0.003) & 0.610 & 1.008 (0.003) \\

\end{tabular}}
    }
    \medskip
    \label{tab:compas}
\end{table}

\subsubsection{Miscellaneous}

\rowcolors{2}{gray!25}{white}
\begin{table}[H]
\centering
\caption{Summary results from the verbal autopsy dataset and Multi-Layer Perceptron (MLP) classifier validated on Dar, Tanzania holdout data with inflated argmax $\epsilon = 0.05$. Standard errors are shown in parentheses and arrow direction denotes higher/lower values are desired. This example demonstrates how the chosen instability definition can alter the effect of the inflated argmax, with the Jaccard definition being noticeably more conservative.}
\resizebox{\textwidth}{!}{%
    {\renewcommand{\arraystretch}{1.5} 
    \begin{tabular}{c c c c c c c c c}
    \rowcolor{gray!60}
    & & \multicolumn{3}{c}{Mean Instability $\searrow$} & \multicolumn{3}{c}{Max Instability $\searrow$} & \\
    \rowcolor{gray!60}
    Sel. rule & Algo & Disjoint & Overlap & Jaccard & Disjoint & Overlap & Jaccard & $\beta_{\text{size}}\searrow$ \\\hline
    \rowcolor{white}
     argmax & $\mathcal{A}$ & 0.155 (0.006) & 0.155 (0.006) & 0.155 (0.006) & 0.986 & 0.986 & 0.986 & 1.000 (0.000) \\
    \rowcolor{gray!25}
     argmax & $\widetilde{\mathcal{A}}_m$ & 0.011 (0.002) & 0.011 (0.002) & 0.011 (0.002) & 0.908 & 0.908 & 0.908 & 1.000 (0.000) \\
    \rowcolor{white}
     argmax$^\varepsilon$ & $\mathcal{A}$ & 0.122 (0.005) & 0.122 (0.005) & 0.156 (0.006) & 0.952 & 0.952 & 0.960 & 1.046 (0.005) \\
    \rowcolor{gray!25}
     argmax$^\varepsilon$ & $\widetilde{\mathcal{A}}_m$ & 0.000 (0.000) & 0.000 (0.000) & 0.012 (0.001) & 0.002 & 0.022 & 0.493 & 1.064 (0.006) \\
    \end{tabular}}
    }
    \medskip
    \label{tab:def-comparison}
\end{table}

\rowcolors{0}{}{}
\begin{table}[H]
\centering
\caption{Sensitivity of Rashomon ensemble instability to the scheme of pruning trees in the Rashomon set with bootstrap perturbations. Lowest values are bolded.}
\begin{tabular}{lcccc}
\toprule
& \multicolumn{2}{c}{HELOC} & \multicolumn{2}{c}{Synthetic} \\
\cmidrule(lr){2-3} \cmidrule(lr){4-5}
Pruning & Mean Instability & Max Instability & Mean Instability & Max Instability \\
\midrule
No Pruning & 0.029 & \textbf{0.568} & 0.191 & 0.938 \\
No Clustering & 0.025 & 0.722 & 0.135 & 0.870 \\
500 Clusters  & - & - & 0.128 & 0.872 \\
100 Clusters  & \textbf{0.018} & 0.690 & 0.136 & 0.886 \\
50 Clusters   & 0.019 & 0.678 & 0.133 & 0.898 \\
25 Clusters   & 0.020 & 0.698 & 0.138 & 0.878 \\
10 Clusters   & 0.023 & 0.630 & \textbf{0.109} & \textbf{0.768} \\
\bottomrule
\end{tabular}
\label{tab:cluster-sensitivity}
\end{table}

\rowcolors{0}{}{}
\begin{table}[H]
\centering
\caption{Mean pairwise disagreement of prediction vectors in the Rashomon ensemble alongside mean and maximum instability, across pruning schemes, on the synthetic dataset with bootstrap perturbations. The highest mean pairwise disagreement and lowest instability measures are bolded. }
\begin{tabular}{lccc}
\toprule
Pruning & Mean Pairwise Disagreement & Mean Instability & Max Instability \\
\midrule
No Pruning     & 0.2181 & 0.191 & 0.938 \\
No Clustering  & 0.2346 & 0.135 & 0.870 \\
500 Clusters   & \textbf{0.2625} & 0.128 & 0.872 \\
100 Clusters   & 0.2520 & 0.136 & 0.886 \\
50 Clusters    & 0.2495 & 0.133 & 0.898 \\
25 Clusters    & 0.2366 & 0.138 & 0.878 \\
10 Clusters    & 0.2324 & \textbf{0.109} & \textbf{0.768} \\
\bottomrule
\end{tabular}
\label{tab:disagree1}
\end{table}

\rowcolors{0}{}{}
\begin{table}[H]
\centering
\caption{Mean pairwise disagreement (PwD) of prediction vectors in the Rashomon set, alongside mean and maximum instability with and without pruning, across SORTeD parameters, on the synthetic dataset with leave-25-out perturbations.}
\begin{tabular}{cccccc}
\toprule
& & \multicolumn{2}{c}{Without Pruning} & \multicolumn{2}{c}{With Pruning} \\
\cmidrule(lr){3-4} \cmidrule(lr){5-6}
\shortstack{Cost\\Complexity} & \shortstack{Mean PwD\\of Rashomon Set} & \shortstack{Mean\\Instability} & \shortstack{Max\\Instability} & \shortstack{Mean\\Instability} & \shortstack{Max\\Instability} \\
\midrule
0.005 & 0.2409 & 0.167 & 0.785 & 0.151 & 0.960 \\
0.01  & 0.2181 & 0.144 & 0.825 & 0.080 & 0.690 \\
\bottomrule
\end{tabular}
\label{tab:disagree2}
\end{table}

\subsection{Synthetic Dataset}
\label{app:synthetic}

The dataset consists of $n = 300$ samples and $p = 15$ features generated via \texttt{make\_classification} with 4 informative features, 6 redundant features (linearly correlated with the informative features), and 1 repeated feature. The outcome is binary with class weights $[0.85, 0.15]$, class separability \texttt{class\_sep} $= 0.25$, and label noise of 10\% applied via random flips. Features are standardized to zero mean and unit variance.

To expand the Rashomon set, the training set is pruned to remove samples near the decision boundary. A logistic regression model $\hat{f}$ is fit on the raw training data, and any sample with predicted probability within 0.2 of the boundary is discarded:

$$\mathcal{D}_{\text{train}} = \left\{(X_i, Y_i) : \left|\hat{f}(X_i) - 0.5\right| > 
0.2\right\}.$$

The test set is augmented with $n_{\text{ambig}} = 500$ synthetic points drawn uniformly from $[-1, 1]^p$ and labeled by the decision function of $\hat{f}$:

$$X_i^{\text{ambig}} \overset{iid}{\sim} \text{Uniform}([-1,1]^p), \quad 
Y_i^{\text{ambig}} = \mathbf{1}\{\hat{f}(X_i^{\text{ambig}}) > 0.5\}.$$

\subsection{Model Configuration \& Hyperparameters}
\label{app:models}

We evaluated these methods with the following models. Most, with the exception of InSilicoVA and SORTeD, are imported from \verb|sklearn|.

\begin{flushleft}
\begin{itemize}[leftmargin=*, itemsep=10pt, topsep=6pt]

\item \textbf{MLP} --- \texttt{MLPClassifier} (preprocessing: \texttt{StandardScaler}) \\
\small\texttt{hidden\_layer\_sizes=(16,), solver='adam', learning\_rate\_init=0.01, alpha=0.01, batch\_size=256, max\_iter=30, early\_stopping=True, n\_iter\_no\_change=3, random\_state=42}\normalsize

\item \textbf{Gradient Boosting} --- \texttt{GradientBoostingClassifier} \\
\small\texttt{n\_estimators=50, learning\_rate=0.05, max\_depth=2, subsample=0.8, max\_features='sqrt', min\_samples\_split=20, min\_samples\_leaf=10, random\_state=1}\normalsize

\item \textbf{Logistic Regression} --- \texttt{LogisticRegression} (preprocessing: \texttt{StandardScaler}) \\
\small\texttt{penalty='l2', C=0.01, solver='saga', tol=1e-3, max\_iter=500, random\_state=0, n\_jobs=-1}\normalsize

\item \textbf{LDA} --- \texttt{LinearDiscriminantAnalysis} (preprocessing: \texttt{StandardScaler}) \\
\small\texttt{solver='lsqr', shrinkage='auto'}\normalsize

\item \textbf{Naive Bayes} --- \texttt{GaussianNB} \\
\small Default settings\normalsize

\item \textbf{KNN} --- \texttt{KNeighborsClassifier} (preprocessing: \texttt{StandardScaler}) \\
\small\texttt{n\_neighbors=5, weights='distance'}\normalsize

\item \textbf{Decision Tree} --- \texttt{DecisionTreeClassifier} \\
\small\texttt{max\_depth=10, min\_samples\_split=5, min\_samples\_leaf=3, max\_features=None, random\_state=0}\normalsize

\item \textbf{Regularized Decision Tree} --- \texttt{DecisionTreeClassifier} \\
\small\texttt{max\_depth=5, min\_samples\_split=20, min\_samples\_leaf=10, random\_state=0}\normalsize

\item \textbf{InSilicoVA} --- \texttt{InSilicoVA} \\
\small\texttt{data.type='PHMRC', phmrc.type='adult', convert.type='fixed', Nsim=5000, auto.length=FALSE}\normalsize

\item \textbf{SORTeD} --- \texttt{SORTDClassifier} \\
\small\texttt{max\_num\_trees=5000, cost\_complexity=0.01 (synthetic), 0.005 (HELOC)}\normalsize

\end{itemize}
\end{flushleft}

\end{document}